\documentclass[journal]{IEEEtran}
\usepackage[usenames,dvipsnames]{color}
\usepackage{colortbl}
\definecolor{mygray}{gray}{.9}
\usepackage{amsmath}
\usepackage{algorithm}
\usepackage{algorithmic}
\usepackage{graphicx}
\usepackage{verbatim}
\usepackage{multirow}
\usepackage{tabularx}
\usepackage{xcolor}
\usepackage{array}

\usepackage{amssymb}
\usepackage{makecell}
\usepackage{subfigure}
\usepackage{diagbox}
\usepackage{rotating}
\usepackage{booktabs}
\usepackage{tablefootnote}

\usepackage{cite}
\usepackage[colorlinks,
            linkcolor=red,
            anchorcolor=blue,
            citecolor=green
            ]{hyperref}

\newcommand{\addFig}[1]{}
\newcommand{\addFigs}[1]{}

\usepackage{subfigure}

\usepackage{balance}
\usepackage{cleveref}
\crefformat{section}{\S#2#1#3} 
\crefformat{subsection}{\S#2#1#3}
\crefformat{subsubsection}{\S#2#1#3}

\ifCLASSINFOpdf
\else
\fi

\begin{document}
%
\title{Salient Object Detection in RGB-D Videos}
%
%
%
\author{Ao Mou, Yukang Lu, Jiahao He, Dingyao Min, Keren Fu$^{*}$, Qijun Zhao \thanks{A. Mou, Y. Lu, J. He, D. Min, K. Fu and Q. Zhao are with the College of Computer Science, Sichuan University, China.\\
\indent K. Fu and Q. Zhao are also with the National Key Laboratory of Fundamental Science on Synthetic Vision, Sichuan University, China.\\
\indent Corresponding author: Keren Fu (Email: fkrsuper@scu.edu.cn).\\
\indent A preliminary version of this work has appeared in \cite{DCTNet}.
}
}

\markboth{JOURNAL OF LATEX CLASS FILES, VOL. 14, NO. 8, AUGUST 2015}%
{Shell \MakeLowercase{\textit{et al.}}: Bare Demo of IEEEtran.cls for IEEE Journals}

\maketitle

\begin{abstract}
Given the widespread adoption of depth-sensing acquisition devices, RGB-D videos and related data/media have gained considerable traction in various aspects of daily life. Consequently, conducting salient object detection (SOD) in RGB-D videos presents a highly promising and evolving avenue. Despite the potential of this area, SOD in RGB-D videos remains somewhat under-explored, with RGB-D SOD and video SOD (VSOD) traditionally studied in isolation. To explore this emerging field, this paper makes two primary contributions: the dataset and the model. On one front, we construct the RDVS dataset, a new RGB-D VSOD dataset with realistic depth and characterized by its diversity of scenes and rigorous frame-by-frame annotations. We validate the dataset through comprehensive attribute and object-oriented analyses, and provide training and testing splits. Moreover, we introduce DCTNet+, a three-stream network tailored for RGB-D VSOD, with an emphasis on RGB modality and treats depth and optical flow as auxiliary modalities. In pursuit of effective feature enhancement, refinement, and fusion for precise final prediction, we propose two modules: the multi-modal attention module (MAM) and the refinement fusion module (RFM). To enhance interaction and fusion within RFM, we design a universal interaction module (UIM) and then integrate holistic multi-modal attentive paths (HMAPs) for refining multi-modal low-level features before reaching RFMs. Comprehensive experiments, conducted on pseudo RGB-D video datasets alongside our proposed RDVS, highlight the superiority of DCTNet+ over 17 VSOD models and 14 RGB-D SOD models. Additionally, insightful ablation experiments were performed on both pseudo and realistic RGB-D video datasets to demonstrate the advantages of individual modules as well as the necessity of introducing realistic depth into VSOD. Our code together with RDVS dataset will be available at \href{https://github.com/kerenfu/RDVS}{https://github.com/kerenfu/RDVS/}.
\end{abstract}


\IEEEpeerreviewmaketitle

\section{INTRODUCTION}

\begin{figure}
	\centering
	\includegraphics[width=0.45\textwidth]{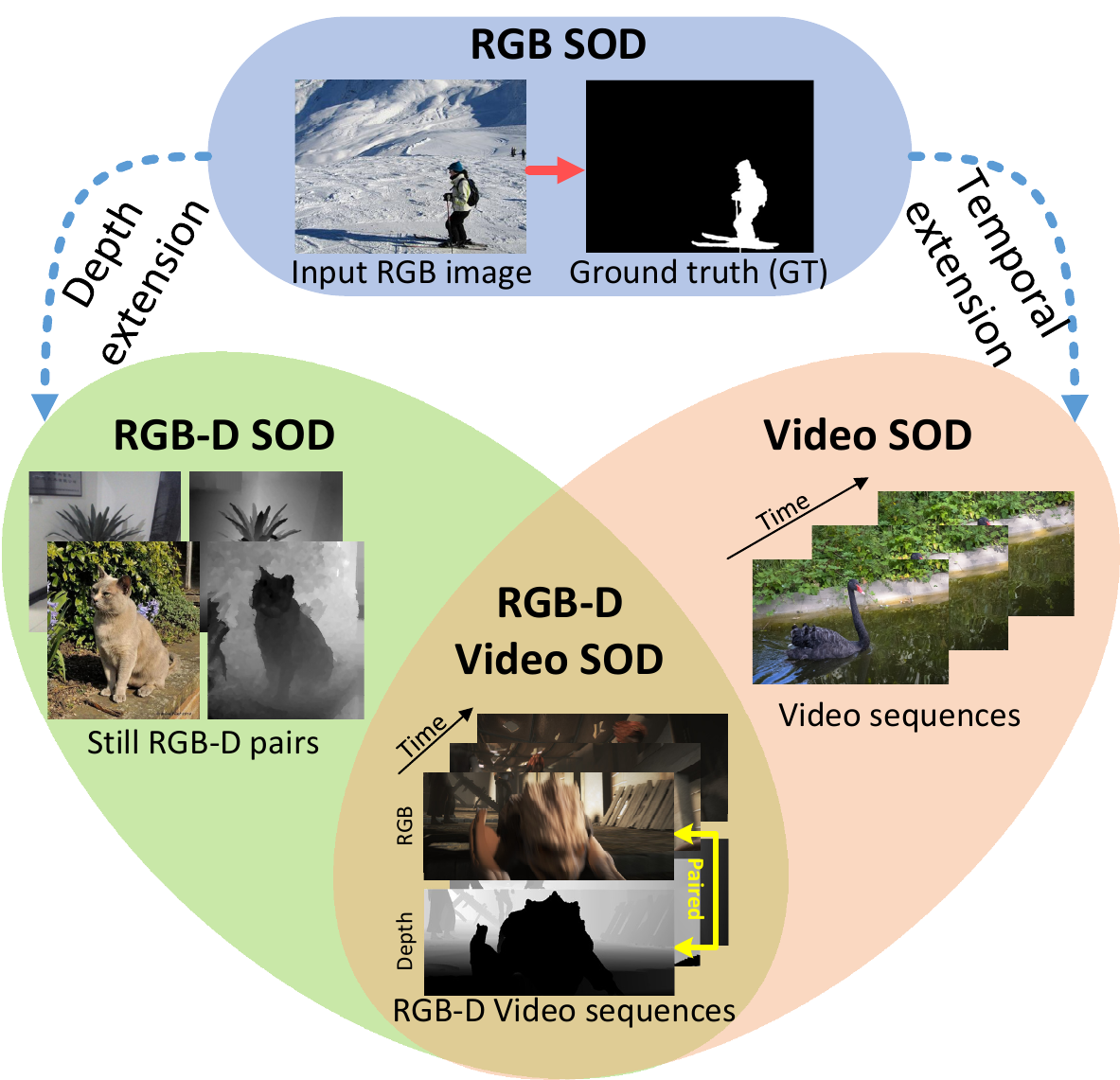}\vspace{-0.1cm}
	\caption{\small We target at the RGB-D VSOD task, which can be deemed as extension from the prevalent RGB-D SOD and VSOD tasks.}\vspace{-0.5cm}
	\label{fig_nodepth}
\end{figure}

\IEEEPARstart{S}{alient} object detection (SOD) aims at locating and segmenting the most attention-grabbing objects in a scene and plays an important role in many advanced computer tasks\cite{wang2021rgbsurvey}, such as object recognition/detection, object segmentation, image retargeting, image manipulation, image retrieval, video tracking and so on. Due to the limitation of using a single RGB/color modality (image) for SOD (termed RGB SOD)\cite{wang2021rgbsurvey}, researchers have integrated scene depth information into the SOD task, often referred to as RGB-D SOD\cite{zhou2021rgbdsurvey}, as illustrated in Fig.\:\ref{fig_nodepth}. Meanwhile, extending still images to the temporal case yields the video SOD (VSOD) task\cite{fan2019shifting}. In the modern deep learning era, RGB-D SOD and VSOD based on convolutional neural networks (CNNs) have gained much attention and grown very rapidly\cite{fu2020jl-dcf,fan2020bbs,li2021hierarchical,fan2019shifting,li2019mga,ji2021fsnet}.

In light of the advances in acquisition devices equipped with depth sensors (\emph{e.g.}, Kinect, Orbbec Astra, LiDAR, and stereo cameras), 
RGB-D videos and related data/media have become increasingly popular in people's life, such as in the scenarios of autonomous driving, robotics, medical imaging, gaming, 3D/IMAX movies, and even daily Vlog/clip recording. Therefore, conducting SOD in RGB-D videos holds significant potential and represents an emerging promising direction.
However, in the saliency community, SOD in RGB-D videos (called RGB-D VSOD hereafter) remains somewhat under-explored, while RGB-D SOD and VSOD are usually investigated separately and independently along their own pathways \cite{zhou2021rgbdsurvey,fan2019shifting,fu2020jl-dcf,fan2020bbs,li2021hierarchical,li2019mga,ji2021fsnet}. Furthermore, the community still lacks a high-quality public dataset which is tailored for this task, hence hindering researchers from probing into this practical and valuable research problem.

RGB-D VSOD can be deemed as temporal and depth extension from RGB-D SOD and VSOD tasks (Fig.\:\ref{fig_nodepth}), respectively. To delve into such a potential task, and as one of the earliest works towards RGB-D VSOD, this paper contributes on two distinct aspects: 1) the dataset, and 2) the model.
On one hand, to establish a dataset that enables RGB-D VSOD, we propose a new RGB-D Video Salient Object Dataset incorporating realistic depth information, and the dataset is named RDVS for short. RDVS contains 57 sequences, totaling 4,087 frames, and its annotation process is guided rigorously by gaze data captured from a professional eye-tracker. Such an annotation process is considered more scientific than naive labelling\cite{fan2019shifting}. The collected video clips encompass various challenging scenarios, \emph{e.g.}, complex backgrounds, low contrast, occlusion, and heterogeneous objects.
It is worth noting that as a follow-up work after our previous conference work \cite{DCTNet}, Zhang \emph{et al.} \cite{zhang2023salient} also noticed this area and collected an RGB-D VSOD dataset, which however has very limited size and scene diversity (20 clips, $\sim$1k frames, and most target objects are human), and their annotation is yet not strictly guided by eye tracker. So, a high-quality RGB-D VSOD dataset is still in the urgent need and would benefit the community a lot.

On the other hand, to achieve accurate SOD in RGB-D videos, we propose a model called \emph{DCTNet+}, which is an improved version from our previous DCTNet \cite{DCTNet}. DCTNet and DCTNet+ are featured as a three-stream network that learns to integrate depth, RGB, and optical flow (OF), which are fed to the network as trimodal inputs. Unlike prior multi-modal approaches \cite{ji2021fsnet,fu2020jl-dcf,zhou2021specificity} that treat different modalities equally and overlook their inherent properties, we draw inspiration from \cite{li2019mga} and allocate one primary modality (\emph{i.e.} RGB) and two auxiliary modalities (\emph{i.e.} depth and OF). Under this setup, we introduce a multi-modal attention module (MAM) and a refinement fusion module (RFM). MAM enriches features by modeling multi-modal long-range dependencies, whereas RFM attenuates noise and refines features through interactive fusion of attention maps between primary and auxiliary modalities, and finally culminates in progressive fusion to achieve fused features for prediction. To benefit universal interaction and fusion within RFM and acquire diversified feature
characteristics, we design a universal interaction module (UIM), which is then broadly utilized in RFM.
To further improve performance, we also propose holistic multi-modal attentive paths (HMAPs) to purify multi-modal low-level features prior to RFMs, which is achieved by refining low-level features using the coarse prediction derived from high-level features.

In a nutshell, this paper provides four main contributions:

\begin{itemize}
\item We construct a new RGB-D VSOD dataset, namely RDVS, containing diverse scenes with scientific frame-by-frame annotations. We validate the dataset with thorough analyses including attribute and object-oriented analyses, and also provide training and testing splits. We believe RDVS could benefit the community and bring insight into related tasks. RDVS dataset together with the benchmark results are available at \href{https://github.com/kerenfu/RDVS}{https://github.com/kerenfu/RDVS/}.

\item We propose DCTNet+, a three-stream network for addressing RGB-D VSOD, which allocates RGB as primary modality, and depth/OF as auxiliary modalities. Two novel modules, namely multi-modal attention module (MAM) and refinement fusion module (RFM), are proposed to achieve effective enhancement, refinement, and fusion of features, leading to accurate final prediction.

\item To benefit interaction and fusion within RFM, we design a universal interaction module (UIM), and further incorporate holistic multi-modal attentive paths (HMAPs) to purify multi-modal low-level features prior to RFMs.

\item Extensive experiments on pseudo RGB-D video datasets as well as the proposed RDVS show the superiority of DCTNet+ against 17 VSOD models and 14 RGB-D SOD models. We also conduct insightful ablation experiments on both pseudo and realistic RGB-D video datasets to show the benefit of individual modules as well as incorporating realistic depth into VSOD.
\end{itemize}

The remainder of the paper is organized as follows. Section \ref{sec:related-work} discusses related work on RGB-D SOD, VSOD, and also RGB-D video datasets. Section \ref{dataset} describes the collection, labeling, and analyses of the RDVS dataset. Section \ref{sec:method} describes the proposed DCTNet+ in detail. Experimental results, performance evaluations and comparisons are included in Section \ref{sec:Experiments}. Finally, conclusions are drawn in Section \ref{sec:conclusion}.

\section{RELATED WORK}
\label{sec:related-work}

\subsection{RGB-D Salient Object Detection} \label{RGB-D SOD}

With the development of hardware devices such as 3D cameras, the utilization of depth information in the SOD task has attracted more research attention.
Modern RGB-D SOD methods are mostly deep learning-driven. Since both RGB and depth features are extracted using deep neural network backbones, how to reasonably fuse these two types of features turns to be the key issue. To this end, fusion-based approaches have exhibited commendable achievements, and prevailing methodologies can be broadly classified into three principal categories\cite{fu2021siamese}: early-fusion, late-fusion, and middle-fusion. For the early-fusion strategy, the depth map is usually incorporated as an adjunct channel to the RGB input \cite{qu2017rgbd,huang2018rgbd,zhao2020single,fan2020rethinking}. For example, Zhao \emph{et al.} \cite{zhao2020single} devised a single-stream network with four-channel inputs, enabling early fusion of RGB and depth maps. A depth-enhanced dual attention module was later integrated to facilitate spatial filtering of features.
Unlike early-fusion, the late-fusion scheme first predicts saliency maps of the RGB and depth branches separately, and then fuses such results to obtain the final prediction \cite{shigematsu2017learning,han2017cnns,wang2019Adaptive,ding2019depth,zhu2019pdnet}. AFNet \cite{wang2019Adaptive} employed a two-stream CNN for feature extraction and saliency map prediction from RGB and depth inputs. A saliency fusion module was utilized to learn switching maps and adaptively merge the predicted saliency maps. Zhu \emph{et al.} \cite{zhu2019pdnet} devised a depth sub-network to extract depth information, aiding the primary RGB network. A deconvolution process served as the decoder to generate a saliency map. In late-fusion, since the cross-modal interactions occur only during the result fusion stage, full exploitation of complementarities between the RGB and depth modalities remain intractable. As a result, the late fusion scheme remains ineffective in discovering complex intrinsic correlations across the modalities.


The middle-fusion scheme, instead, conducts separate extraction of RGB and depth features, followed by feature fusion at various stages to achieve the final prediction. This strategy has become a prevailing manner in the field of RGB-D SOD. Representative methods include the bifurcated backbone approach \cite{fan2020bbs}, depth potential estimation \cite{chen2020dpanet}, depth-calibrated fusion \cite{ji2021calibrated}, bidirectional encoder interactions \cite{zhang2021bts}, two-stage cross-modal fusion \cite{jin2021cdnet},  adaptive feature re-weighting \cite{li2021hierarchical}, Siamese network \cite{fu2020jl-dcf}, etc.
More recently, Wang \emph{et al.} \cite{wang2022learning} introduced a lightweight depth branching method to extract parallel depth features meanwhile preserving the spatial structure. In \cite{chen20223d}, 3D CNNs were used for bimodal pre-fusion, strengthening cross-modal fusion in the encoder stage. Song \emph{et al.} \cite{song2022improving} introduced a modality-aware decoder to enhance multi-scale feature aggregation by considering relations of modalities. Work \cite{pang2023caver} expanded Transformer's attention to both spatial and channel context, and incorporated a convolutional feed-forward network to improve local details.
Within the middle-fusion framework, researchers have also focused on the efficiency issue beyond accuracy. In \cite{wu2021mobilesal}, a MobileNet-V2 backbone was employed and RGB and depth features were fused at the coarsest level. In DFM-Net \cite{zhang2021depth}, the RGB branch employed MobileNet-V2, while the depth branch used a customized super-light backbone network whose channels and layers were reduced. Depth features were also enhanced prior to cross-modal fusion, in order to boost detection accuracy.

Regarding the RGB-D SOD task, the research focus is mostly on optimizing feature fusion, in order to leverage diverse cues, and depth cues have been proved a valuable supplement to the RGB modality. Likewise, in this paper, our work aims at effectively incorporating extra depth information to further boost VSOD accuracy.

\subsection{Video Salient Object Detection} \label{VSOD}

Video salient object detection is challenging since it requires joint consideration of both motion and appearance cues.
Yet, conventional VSOD models struggle in tackling complex video scenarios, and therefore deep learning remains to be predominant in modern VSOD technology. Wang \emph{et al.} \cite{wang2017video} were the first to employ a fully convolutional neural network into this task, yielding encouraging results.
Previous emerging approaches were mainly designed based on recurrent neural networks (RNNs), for example, the optical flow-guided recurrent neural encoder \cite{li2018fgrne}, deep bi-directional ConvLSTM \cite{song2018pdbm}, and ConvLSTM with a saliency shift mechanism \cite{fan2019shifting}. Notably, Fan \emph{et al.} \cite{fan2019shifting} also introduced a densely annotated VSOD dataset called DAVSOD, which provides over twenty thousand annotated frames and is the largest dataset for VSOD up to date. Besides, Chen \emph{et al.} \cite{chen2019improved} introduced long-term spatial-temporal information integration into an existing VSOD framework to alleviate short-term hollow effects and reduce false alarms. To address short-term issues, work \cite{li2019accurate} leveraged spatial-temporal coherence across saliency elements and object priors to acquire meaningful long-term insights.

More recently, Gu \emph{et al.} \cite{gu2020pcsa} created a constrained self-attention mechanism as a substitute for non-localized operations, capturing motion cues and grouping them in a pyramid structure. 
Ren \emph{et al.} \cite{ren2020tenet} introduced a triple excitation network, which enhances object boundaries and spatial-temporal salient regions.
Zhang \emph{et al.} \cite{zhang2021dcfnet} improved model adaptation towards real-world scenarios by dynamically generating context-sensitive convolutional kernels.
Moreover, FSNet \cite{ji2021fsnet} extracted features from appearance and motion branches and constrained them to each other through a full-duplex strategy. 
Li \emph{et al.} \cite{li2019mga} introduced an architecture with appearance and motion branches linked by a bridging motion-guided attention module.
Subsequently, both \cite{chen2021confidence} and \cite{liu2022ds} drew inspiration from \cite{li2019mga} and developed their distinct two-stream encoder-decoder approaches.
As an early work introducing the Transformer into VSOD, Min \emph{et al.} \cite{min2022mgtnet} proposed a Transformer module inserted across CNNs to establish long-range mutual guidance. In addition, Liu \emph{et al.} \cite{liu2023costformer} devised long local and short global Transformers to gather comprehensive spatial-temporal information and introduced optical flow-guided windowed attention for precise long-term spatial-temporal context capture. They also proposed to use static image datasets to generate pseudo-videos for enhanced model training.

While the research community has extensively explored video salient object detection, existing methods primarily emphasize efficient utilization of temporal and spatial information. Recurrent neural networks and optical flow-based approaches are employed to extract temporal cues, but the integration of scene depth data to aid video salient object detection remains notably under-explored. Nevertheless, as discussed in Sec.\:\ref{RGB-D SOD}, utilizing scene depth information to enhance video salient object detection is intuitively viable. To this end, we conduct the earliest systematic research in the field to incorporate extra depth information for VSOD, and propose a trimodal network to tackle SOD in RGB-D videos.

\subsection{RGB-D Video Datasets in the Community} \label{RGB-D-Video}
Up to date, RGB-D video datasets have already benefited many tasks/applications in the computer vision community. Existing RGB-D video datasets are primarily constructed for
tasks such as simultaneous localization and mapping (SLAM) \cite{Ford}, 3D reconstruction \cite{ScanNet}, monocular depth estimation \cite{wang2019web}, object detection \cite{Waymo}, object recognition \cite{RGBD}, semantic
segmentation \cite{silberman2012indoor}, instance segmentation \cite{Wood}, visual odometry \cite{Midair}, tracking \cite{Argov}, action recognition \cite{Msr}, background subtraction \cite{camplani2017benchmarking}, \emph{etc}. Below review some representative RGB-D video datasets, whose depth is realistic, instead of synthetic.

MSR Action3D \cite{Msr} is one of the earliest RGB-D video datasets, focusing on human action recognition.
The Ford Campus Vision and LiDAR dataset \cite{Ford} was a pioneering work for SLAM, providing valuable insight for coming research in this field.
ScanNet \cite{ScanNet} is one of the most popular datasets for 3D reconstruction, annotated with 3D camera poses and surface reconstructions.
The WSVD dataset \cite{wang2019web} consists of stereo videos collected from the Internet and was originally established for monocular depth estimation.
MPI Sintel \cite{butler2012naturalistic} dataset was purposed at optical flow training and evaluation, but it also offers precise scene depth clues.
The RGB-D Object Dataset proposed in \cite{RGBD} is a large dataset initially constructed for object recognition and detection. However, it is limited to constrained indoor scenes.
Subsequently, the Mid-Air \cite{Midair} dataset contains numerous outdoor scenes which enable various vision learning tasks, \emph{e.g.,} visual odometry and SLAM. 
ARKitTrack \cite{Argov} is an emerging dataset for RGB-D tracking, covering both static and dynamic scenes.
Waymo \cite{Waymo} is a large-scale, high quality, and diverse dataset, which facilitates 3D object detection in autonomous driving context.
George \emph{et al.} \cite{Eyefix} constructed an RGB-D video dataset for eye-fixation prediction and therefore video saliency estimation, which is closely related to our work. Unfortunately, the constructed dataset was later not released to the public.
In contrast, the SBM-RGBD dataset \cite{camplani2017benchmarking} includes so-called saliency masks, which however are not manually labeled for SOD. Instead, it aims for background subtraction.
More recently, an RGB-D video dataset aiming at SOD was mentioned in \cite{zhang2023salient}. Although part of videos in this dataset have realistic depth (others have the synthetic one), only $\sim$1k frames with realistic depth are collected. Besides, their ground truth annotation process is not that rigorous.
We refer readers to \cite{lopes2022survey} for a more comprehensive survey on RGB-D video datasets.

Despite the abundance of RGB-D video datasets in the community, there still lacks a well-designed RGB-D video dataset on which SOD can be achieved and evaluated. Thus, we propose a new such dataset and make it publicly available.

\section{Proposed RDVS dataset} \label{dataset}

\begin{table*}[!ht]
    \centering
    \footnotesize
    \caption{Overview of the video selection for the proposed RDVS dataset.}
   \vspace{-5pt}
    \renewcommand{\arraystretch}{1.2}
    \renewcommand{\tabcolsep}{1.1mm}
    \begin{tabular}{c|c|cccc}
        \hline
        Contributing Dataset & Original Usage & Depth Source  &  Selected Sequences & Collected Frames & Resolution\\
        \hline
        MPI Sintel \cite{butler2012naturalistic} & Optical flow evaluation & Computer graphics  &  13 & 569 & 1024$\times$436\\
        \hline
        WSVD \cite{wang2019web} & Monocular depth estimation & Binocular \& disparity  &  29 & 2139 & 1280$\times$720\\
        \hline
       Stereo Ego-Motion \cite{stereego} & Not publicly claimed & Binocular \& disparity  &  10 & 884 & 960$\times$540\\
        \hline
        SBM-RGBD \cite{camplani2017benchmarking} & Moving
object detection & Microsoft Kinect  &  4 & 418 & 640$\times$480\\
        \hline
        TUM-RGBD \cite{sturm2012benchmark} & RGB-D SLAM & Microsoft Kinect &  1 & 77 & 640$\times$480\\
        \hline
    \end{tabular}
    \label{tab_sources}
    \vspace{-15pt}
\end{table*}

\subsection{Data Collection} \label{data collection}

We construct a new RGB-D video dataset for the salient object detection (SOD) task, which includes \emph{realistic depth information} obtained through measurement, rather than being synthetic or estimated from monocular depth estimation. This dataset is the first of its kind in the field and is referred to as the RGB-D Video Salient Object Dataset (RDVS). The RDVS dataset is constructed by combining selected RGB-D videos from five previously publicized datasets, namely MPI Sintel \cite{butler2012naturalistic}, WSVD \cite{wang2019web}, Stereo Ego-Motion \cite{stereego}, SBM-RGBD \cite{camplani2017benchmarking}, and TUM-RGBD \cite{sturm2012benchmark}. While previous literature has proposed various RGB-D video datasets, as reviewed in \cite{lopes2022survey}, they were designed for different visual tasks such as simultaneous localization and mapping (SLAM) \cite{sturm2012benchmark}, tracking \cite{song2013tracking}, or autonomous driving (\emph{e.g.}, the well-known KITTI \cite{geiger2012we}), and they often contained complex/cluttered or even non-salient scenes that were not conducive to identifying unambiguous salient objects. The RDVS dataset was selected and constructed with careful consideration, and Table\:\ref{tab_sources} presents an overview of the video selection for our proposed RDVS dataset. In total, we collect 57 sequences consisting of 4,087 frames, with the depth sources obtained through three distinct means: computer graphics, binocular disparity computation, and real depth sensors. The frame resolution varies from 640$\times$480 to 1280$\times$720. Note that some depth maps of SBM-RGBD, acquired via Kinect structured light sensors, contain null values due to interference. So, we employ the technique described in \cite{silberman2012indoor} for completion.
Although this RDVS dataset may not be exceptionally large, however, it enables the community to investigate this issue under the context of realistic depth, providing interesting findings (\emph{e.g.}, how realistic depth compares with synthetic depth in this task, see Sec.\:\ref{rdvs.3}), shedding light on potential directions for future research.

\subsection{Dataset Annotation} \label{dataset annotation}
The collected RGB-D videos are annotated with salient objects, where pixel-wise binary segmentation of salient objects is desired. Noting that, annotating salient objects in videos is much more difficult than that for still images where participants can just click on segments \cite{li2014secrets} or draw bounding boxes \cite{peng2014rgbd,liu2021learning} of objects per image individually. Since temporal stimuli/consistency should be concerned when people identify salient objects in videos, using eye-tracker has been identified as a more reasonable and feasible way \cite{wang2019revisiting}. Following previous works \cite{wang2019revisiting,fan2019shifting}, we employ an eye-tracker to help our annotation process and identify biologically-inspired salient objects that participates would naturally focus on when watching the videos.


To be specific, we utilize a Tobii Eye Tracker 4C to acquire gaze point data from participants during unconstrained viewing of video stimuli. The Eye Tracker operates at a sampling rate of 90Hz, while the visual stimuli are presented on a 23.8'' screen with a resolution of 1920$\times$1080.
For videos from WSVD and Stereo Ego-Motion, we directly resize their frames to occupy the full screen, while for SBM-RGBD, TUM-RGBD, and MPI Sintel, we try to scale the frames to the full screen size but still retain their aspect ratios, and the remaining screen regions are kept black.
To minimize the impact of head movements, a headrest is employed to maintain participants' heads at a distance of approximately 70 cm from the screen, in accordance with the manufacturer's guidelines.

16 participants (9 males and 7 females, aging between 22 and 35) who passed the eye tracker calibration and have normal or corrected-to-normal vision were involved in our eye tracking experiment. They have not participated in any eye-tracking experiment nor seen these videos before. During the experiment, they are asked to do free-viewing over the videos and are generally naive to the underlying purposes of the experiment. Noting that in our work, we suppose that depth information is only to assist saliency detection as an auxiliary source, but will not be displayed to participants to affect their attention. Therefore, only the RGB frames are displayed to participants as stimuli. This treatment is consistent with the annotation of previous RGB-D SOD datasets such as SIP \cite{fan2020rethinking}, NLPR \cite{peng2014rgbd}, and recent RedWeb-S \cite{liu2021learning}.

\begin{figure*}[t]
    \includegraphics[width=1\textwidth]{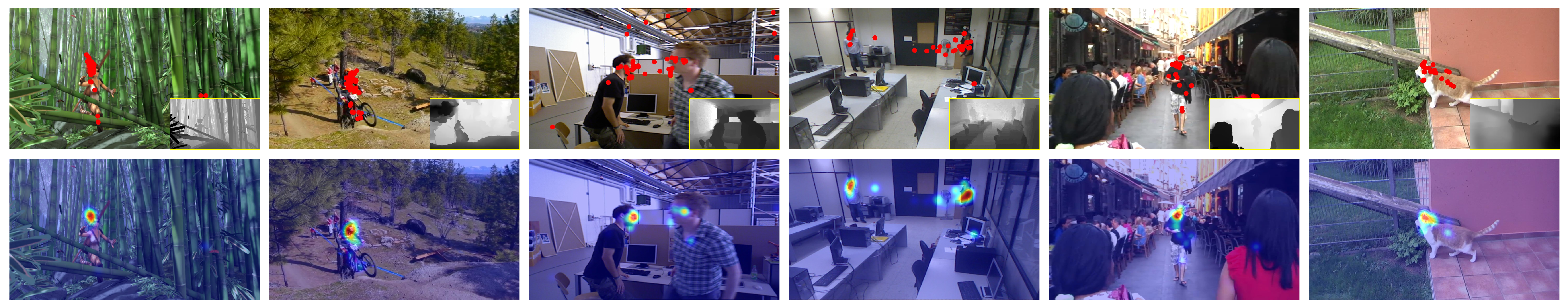} \vspace{-0.8cm}
    \caption{\small Illustrative frames (with depth in the bottom-right) from RDVS with fixations (red dots, the top row) and the corresponding  continuous saliency maps (overlaying on the RGB frames, the bottom row).}
    \label{fig_fixation}
    \vspace{-0.4cm}
\end{figure*}

The collected gaze points (of all participants) in our RDVS dataset are exemplified in Fig.\:\ref{fig_fixation}. Next, following \cite{fan2019shifting,wang2019revisiting}, we first apply a small Gaussian filter to these sparse fixation points to obtain continuous saliency maps, which are then used to guide the subsequent annotation for salient object masks. 10 human annotators, who were pre-trained with several video examples, are instructed to select main objects per-frame according to the corresponding fixation records and meticulously annotate them with fine boundaries instead of rough polygons.
Also note that since we annotate salient objects according to real eye fixations, our dataset is able to present certain \emph{saliency shift}, namely that the video salient object(s) may dynamically change as featured in DAVSOD \cite{fan2019shifting}.

\subsection{Dataset Statistics and Comparisons} \label{dataset statistics}

Finally, our RDVS dataset yields 57 fully annotated video sequences, and totally has 4,087 frames. A straightforward comparison of RDVS with existing commonly used VSOD datasets is shown in Table\:\ref{tab_datasets}, where we can highlight several advantages of RDVS: 1) RDVS is specially constructed for the SOD task. Comparing to DAVIS, FBMS, and SegTrack-V2, the entire annotation process is guided by eye fixations from the eye-tracker, and therefore is more reasonable and scientific. 2) RDVS covers the attention-shift cases (28 sequences out of 57), also called saliency shift as in \cite{fan2019shifting}. Therefore, it is enabled to reflect such an instinctive visual attention behavior of human, which would occur in realistic and dynamic scenes. 3) Every frame in RDVS is annotated with ground truth saliency masks. This contrasts to FBMS and VOS whose frames are not densely labeled. 4) Only RDVS provides per-frame realistic depth maps aligned with the RGB frames.

\begin{table}
    \caption{Statistics and comparisons of previous VSOD datasets and the proposed RDVS dataset. \#Seq.: number of video sequences. \#AF.: number of annotated frames. DL: whether provide densely (per-frame) labeling. AS: whether consider attention shift. FP: whether annotate salient objects according to eye fixation records. RD: whether offer real depth together with RGB frames.}
	\centering
    \renewcommand{\tabcolsep}{1.5mm}
    \begin{tabular}{r|c r r|c c c c}
    \hline
    Dataset & Year & \#Seq. & \#AF. & DL & AS & FP & RD \\
    \hline
    SegTrack-V2\cite{segv2} &2013 & 14 & 1065 & $\checkmark$ &  &  &  \\
    FBMS\cite{fbms} &2014 & 59 & 720 &  &  &  &  \\
    MCL\cite{MCL} &2015 & 9 & 463 &  &  &  &  \\
    ViSal\cite{visal} &2015 & 17 & 193 &  &  &  &  \\
    DAVIS\cite{davis} &2016 & 50 & 3455 & $\checkmark$ &  &  &  \\
        UVSD\cite{UVSD} &2017 & 18 & 3262 & $\checkmark$ &  &  &  \\
    VOS\cite{vos} &2018 & 200 & 7467 &  &  & $\checkmark$ &  \\
    DAVSOD\cite{fan2019shifting} &2019 & 226 & 23938 & $\checkmark$ & $\checkmark$ & $\checkmark$ &  \\
    RDVS (Ours) & -- & 57 & 4087 & $\checkmark$ & $\checkmark$ & $\checkmark$ & $\checkmark$ \\
    \hline
    \end{tabular}
    \label{tab_datasets}
\end{table}

\begin{table}
    \caption{List of video attributes and the corresponding description. We refer to \cite{davis} and extend a part of their attributes (top) with extra four more general attributes (bottom) regarding extreme object sizes and environment.
    }
	\centering
    \begin{tabular}{l|p{0.4\textwidth}<{\raggedright}}
    \hline
    Att. & Description\cr
    \hline
     HO  & \emph{Heterogeneous Object}. Object regions have distinct colors.\\
     OCC & \emph{Occlusion}. Object becomes partially or fully occluded. \\
     OV  & \emph{Out-of-view}. Object is partially clipped by the image boundaries. \\
     FM  & \emph{Fast-Motion}. The average per-frame object motion, computed as centroids Euclidean distance, is larger than 20 pixels.\\
     MB  & \emph{Motion Blur}. Object has fuzzy boundaries due to fast motion. \\
     DEF & \emph{Deformation}. Object presents complex non-rigid deformations.\\
     SC  & \emph{Shape Complexity}. Object has complex boundaries such as thin parts and holes.\\
     SV  & \emph{Scale-Variation}. The area ratio among any pair of bounding boxes enclosing the target object is smaller than 0.5.\\
     AC  & \emph{Appearance Change}. Noticeable appearance variation, due to illumination changes and relative camera-object rotation.\\
     BC  &  \emph{Background Clutter}. The background and foreground regions around object boundaries have similar colors.\\
    \hline
     BO  &	\emph{Big Object}. Ratio of object and image areas becomes larger than 0.25\\
     SO  &	\emph{Small Object}. Ratio of object and image areas becomes less than 0.01\\
     IN  &	\emph{Indoor Scenes}. Objects are captured in indoor environment.\\
     OUT &	\emph{Outdoor Scenes}. Objects are captured in outdoor environment.\\
    \hline
    \end{tabular}
    \label{tab_attribute}
\end{table}

\begin{figure}
    \includegraphics[width=.48\textwidth]{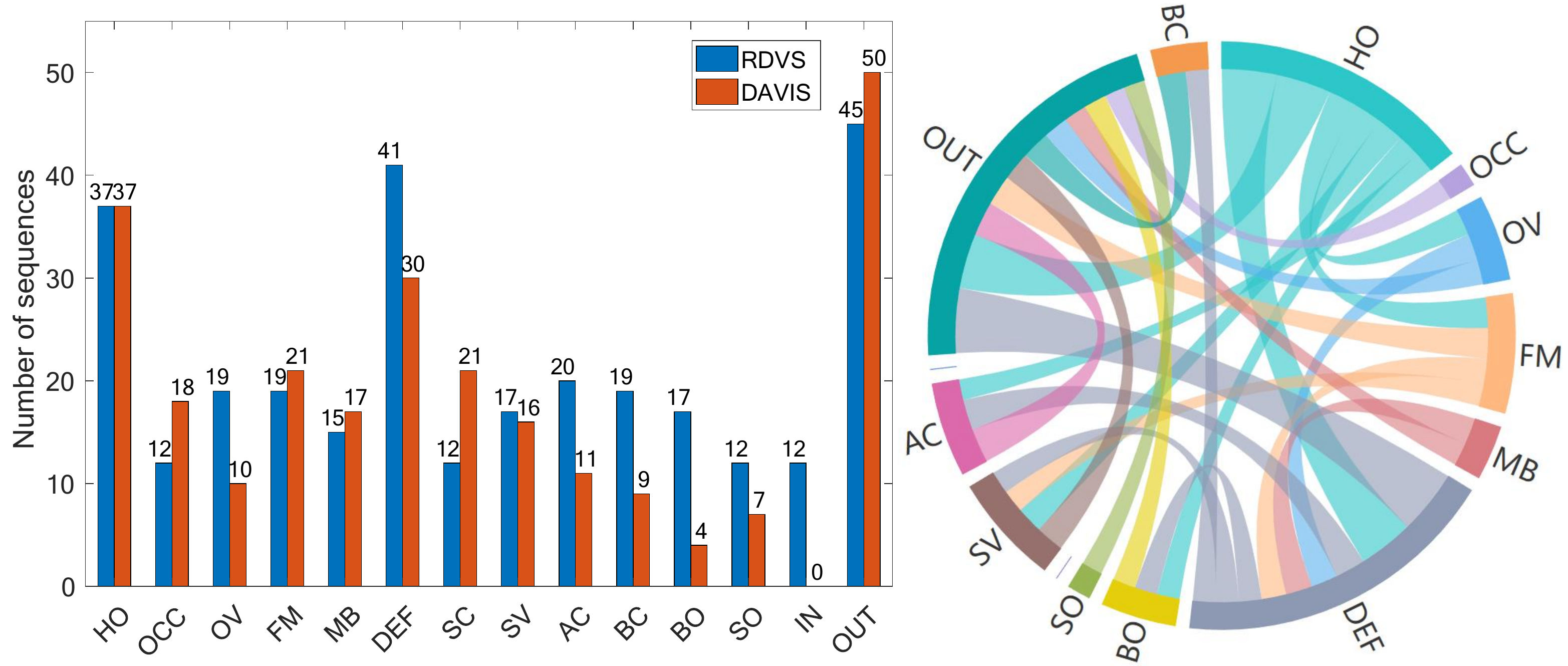}
    \vspace{-0.3cm}
    \caption{\small Attribute-based analyses of RDVS with comparison to DAVIS (left), and the pairwise dependencies across different attributes (right).}
    \label{fig_attributes}
    \vspace{-0.4cm}
\end{figure}

\textbf{Attribute Analyses of RDVS.} To reflect real-world scenarios, a good dataset is supposed to cover a variety of challenging scenes. To verify this, we refer to DAVIS \cite{davis} and summarize the following 14 attributes for RDVS, as detailed in Table\:\ref{tab_attribute}. Compared to DAVIS, we keep the 10 out of 15 attributes, namely discarding EA (Edge Ambiguity), LR (Low Resolution), DB (Dynamic Background), IO (Interacting Objects), CS (Camera-Shake), as those attributes are less applicable to RDVS. Additionally, four more attributes are added: BO (Big Object), SO (Small Object), IN (Indoor), and OUT (Outdoor).
Note most of these attributes are not exclusive, so a sequence can be annotated with multiple attributes.
The distribution of these attributes over the entire RDVS dataset and their pairwise dependencies are shown in Fig.\:\ref{fig_attributes}, with comparison to DAVIS. From Fig.\:\ref{fig_attributes}, one can see that compared with the universally agreed DAVIS dataset, our RDVS has comparative attribute distribution and is even superior to DAVIS on the quantities of OV, DEF, AC and BC attributes, where about 10 more sequences have been achieved. This demonstrates that RDVS is equally or even more challenging than DAVIS. Meanwhile, as attributes IN and OUT are exclusive, we can observe in Fig.\:\ref{fig_attributes} that in RDVS there are a few indoor scenes besides outdoor ones. As a result, the RDVS dataset is more representative of open-world scenarios, providing a fair reflection of real-world situations.

\begin{figure}
    \includegraphics[width=.45\textwidth]{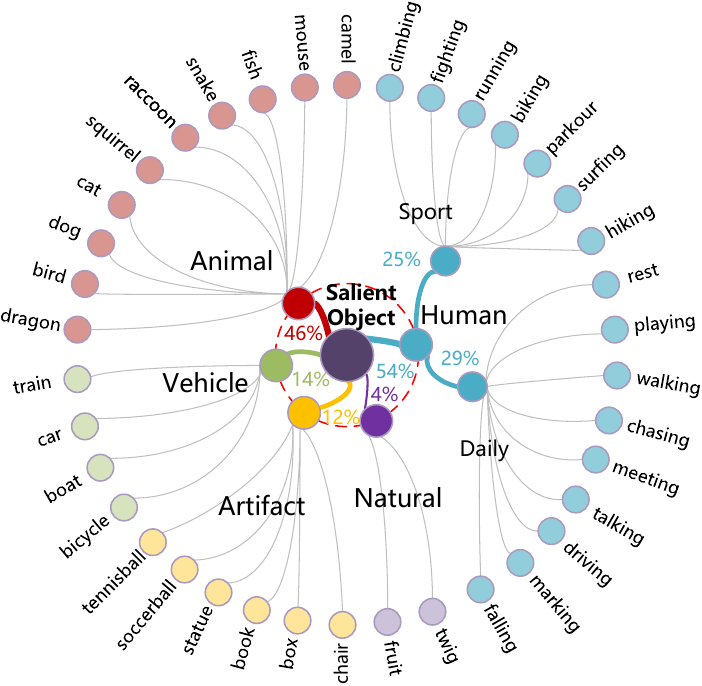}
    \vspace{-0.3cm}
    \caption{\small Scene/object categories of RDVS.}
    \label{fig_categories}
    \vspace{-0.4cm}
\end{figure}

\textbf{Object-oriented Analyses of RDVS.} To conduct object-oriented analyses of RDVS in order to provide deeper understanding, we first summarize the object categories from RDVS in Fig.\:\ref{fig_categories}. Consistent with \cite{fan2019shifting,wang2019revisiting}, a sequence is manually labeled with one or more category (\emph{i.e.}, Animal, Vehicle, Artifact, Human Activity, and Natural). Note we add a ``Natural'' category as a small amount of objects (\emph{e.g.}, a twig bitten by a dog, or a fruit held by a man) cannot be categorized into the former four. Also note these category labels are not exclusive, so the total summation of proportion is larger than 100\%. Furthermore, Human Activity has two exclusive sub-classes: Sports and Daily. As a result, we have built a list of about 38 most frequently present scenes/objects, with their distribution shown in Fig.\:\ref{fig_categories}. One can see that the objects in RDVS span a diverse set of classes, including human activities (\emph{e.g.}, hiking, biking, surfing), artifact (\emph{e.g.}, book, chair, football), vehicle (\emph{e.g.}, train, bicycle, car), and animal (fish, bird, cat, dog). This leads to a relatively comprehensive coverage of object-level saliency in daily-life scenes.

\begin{table}
    \caption{Statistics regarding object instance numbers and motion patterns in RDVS.}
	\centering
	\renewcommand{\tabcolsep}{2.5mm}
    \begin{tabular}{c|ccc|ccc}
    \toprule
    \multirow{2}{*}{RDVS}&\multicolumn{3}{c|}{\# Salient Objects}&
    \multicolumn{3}{c}{Object Motion}\cr
    &1&2&$\geq3$&Slow&Fast&Stable\cr
    \midrule
    \# Videos& 26 & 25& 6& 29&  19&9\cr
    \bottomrule
    \end{tabular}
    \label{tab_motion}
\end{table}

We also show statistics regarding object numbers and motion patterns in RDVS in Table\:\ref{tab_motion}. Accordingly, one can see that most video sequences in RDVS present one or two objects, whereas sequences having more than two objects occupy a small amount, \emph{i.e.}, only six sequences. For motion patterns, sequences with slow motion takes the majority of RDVS, while sequences with fast/no (stable) motion  are fewer. This differs from DAVSOD \cite{fan2019shifting} where over 51\% video clips of DAVSOD have stable objects.

\begin{figure}
    \includegraphics[width=.45\textwidth]{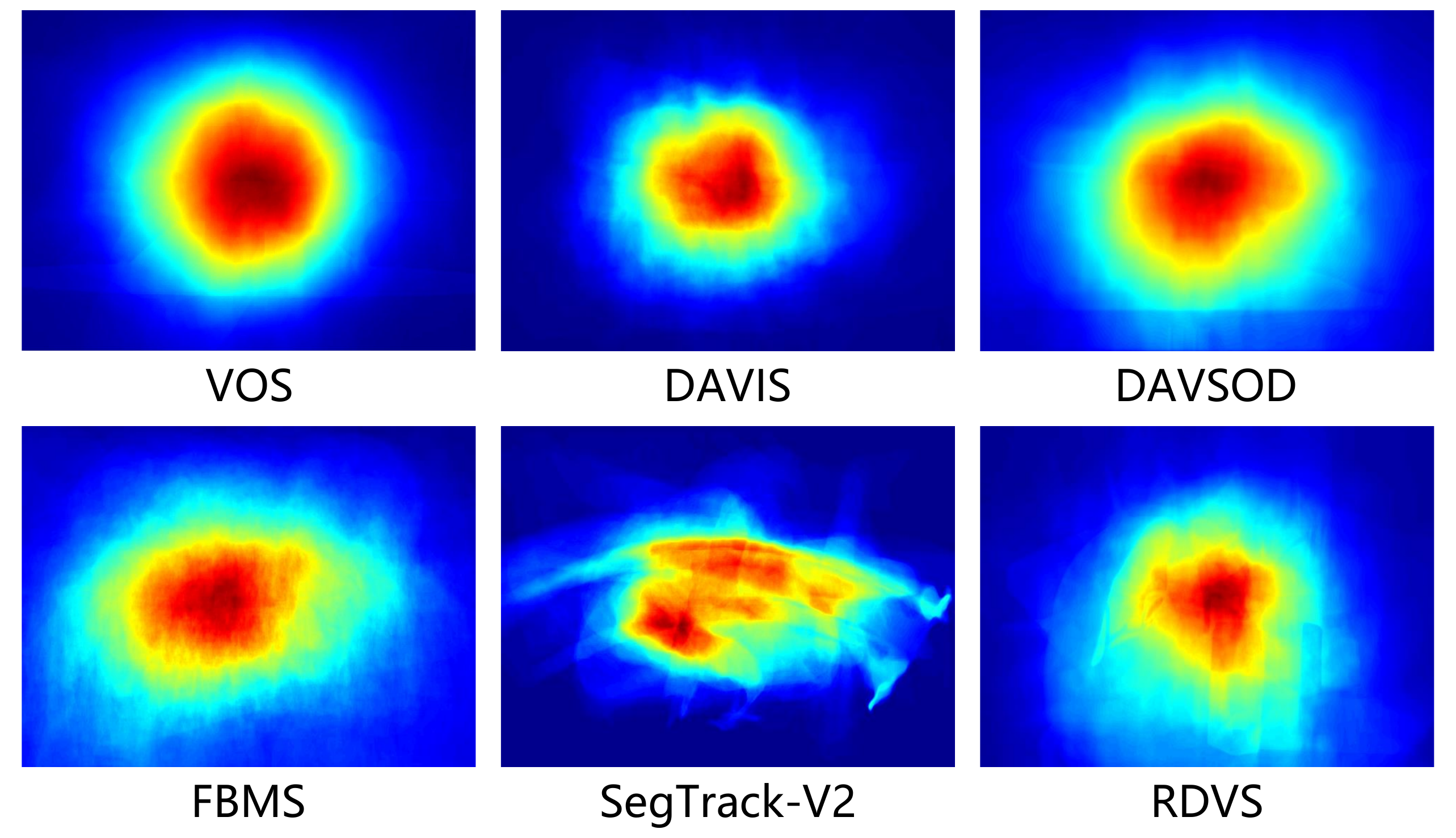}
    \vspace{-0.3cm}
    \caption{\small Center bias of RDVS and existing VSOD datasets.}
    \label{fig_center_bias}
    \vspace{-0.4cm}
\end{figure}

Fig.\:\ref{fig_center_bias} shows the average saliency map over all frames of RDVS together with those of other existing video SOD datasets. As human are more likely to pay attention to the central area of an image/scene during free viewing, one can see the average saliency map of RDVS honestly shows certain degree of center bias, similar to other datasets, though some differences can be observed. Also note that a larger number of video frames tend to yield a smoother average map, otherwise the result may be affected by individual object masks and become less smoother, like the one of SegTrack-V2 in Fig.\:\ref{fig_center_bias}.

\begin{figure}
    \includegraphics[width=.48\textwidth]{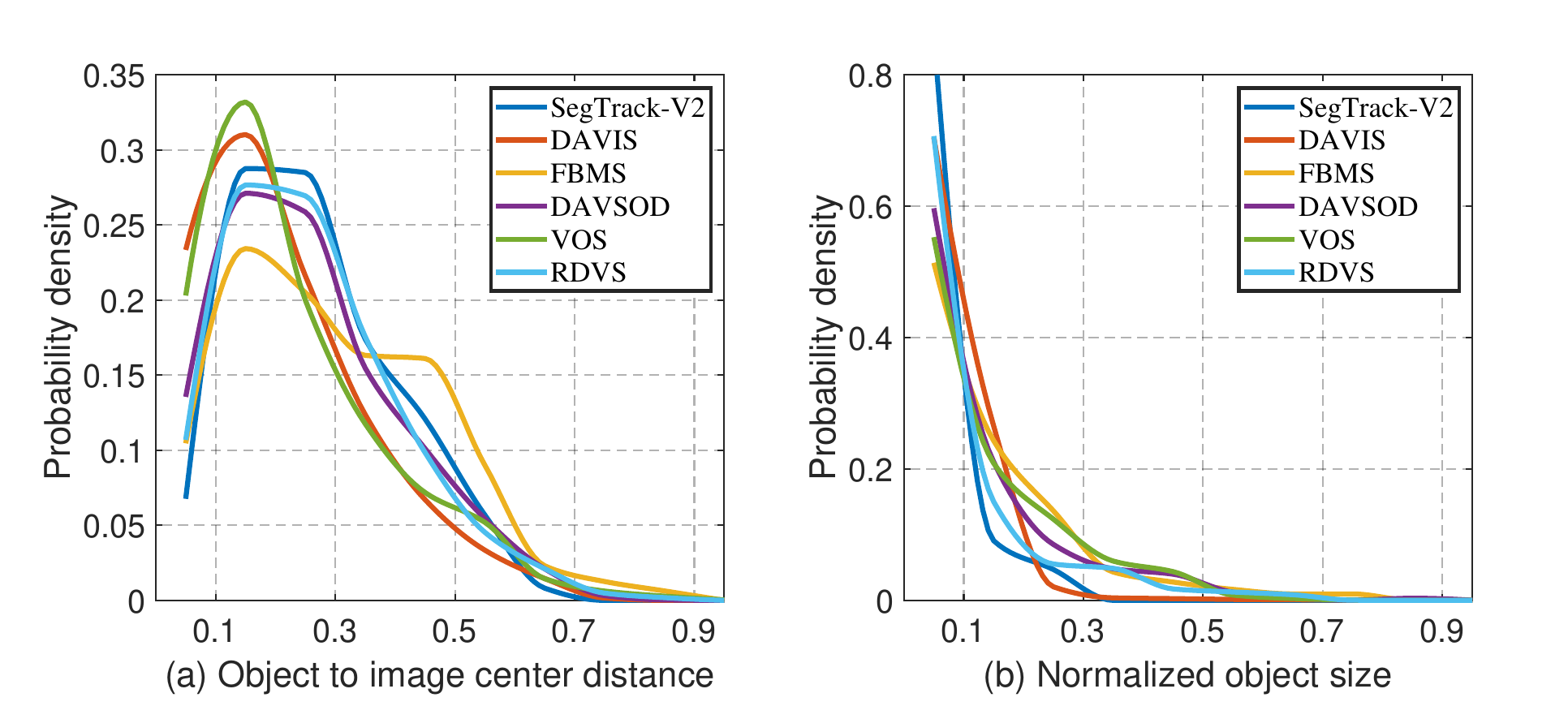}
    \vspace{-0.3cm}
    \caption{\small Statistical summaries of RDVS and existing VSOD datasets, including: (a) distributions of normalized distance between object and image center, and (b) distributions of normalized object size.}
    \label{fig_distribution}
    \vspace{-0.4cm}
\end{figure}

Since the above average saliency map only provides a general intuition, to better understand the cases for different objects, we conduct two statistical analyses, including distributions of normalized object distances from image centers, as well as size ratios of salient objects. In this statistics, frames in a sequence are treated independently. Results are shown in Fig.\:\ref{fig_distribution}, from which we can see all datasets present certain center bias, while objects from VOS are generally the closest to the image centers. In contrast, the probability distribution of RDVS is similar to DAVSOD. Fig.\:\ref{fig_distribution} (b) shows that object sizes of all datasets vary smoothly from small to large, and most objects have size ratios lower than 0.5. Overall, our RDVS has a relatively larger proportion of small objects.

In all, through the above object-oriented analyses of RDVS, one can see that the statistics of RDVS resemble existing public video salient object detection datasets. This proves that our RDVS dataset is a reasonable and feasible video salient object detection with realistic depth.

\subsection{Dataset Training/Testing Splits} \label{dataset splits}
We setup two training and testing protocols for RDVS. The first one is to use the entire RDVS for testing, which aims to test the generalizability of algorithms trained on other datasets/sources on the proposed RDVS.
The second one is to divide RDVS into the default training and testing sets. The training set comprises 32 sequences, amounting to 2,208 frames, while the test set consists of 27 sequences, totaling 1,879 frames. The division of the training and test sets has ensured balance across object categories and attributes.


\begin{figure*}
    \includegraphics[width=1.0\textwidth]{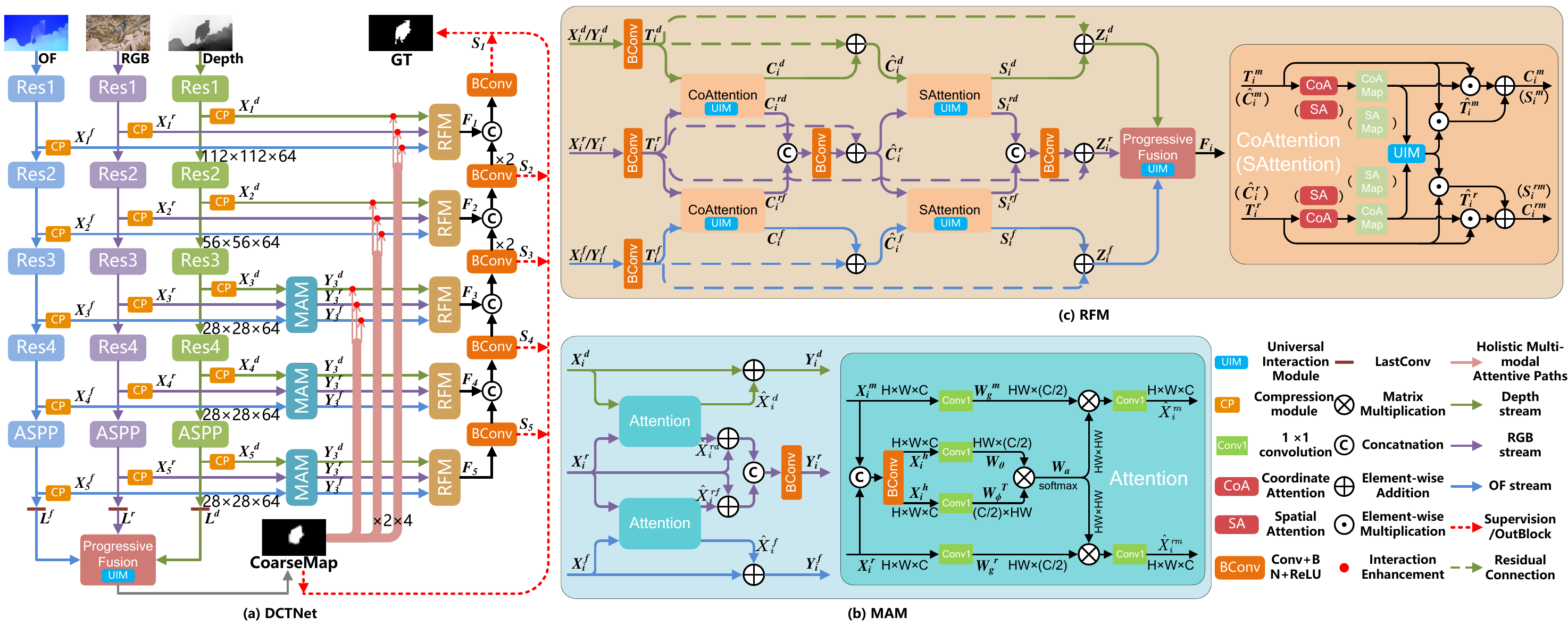}
    \vspace{-0.6cm}
    \caption{\small Overview of DCTNet+. (a) shows the big picture. (b) and (c) show the details of MAM and RFM, respectively. In the attention operations on the right-hand side in (c), since the coordinate attention and spatial attention processes are similar, the operations of spatial attention are represented in parentheses and are not repeated.}
    \label{fig_blockdiagram}
    \vspace{-0.4cm}
\end{figure*}

\section{METHODOLOGY}
\label{sec:method}
The cope with RGB-D VSOD, one needs to consider three key factors: spatial, temporal, and depth. We propose DCTNet+, whose framework is shown in Fig.\:\ref{fig_blockdiagram} and is characterized by a three-stream encoder-decoder architecture. Optical flow (OF) maps are rendered by RAFT \cite{raft}. Each stream of the encoder is based on ResNet-34 \cite{resnet}, and following \cite{li2019mga}, the ASPP (Atrous Spatial Pyramid Pooling \cite{deeplab}) module is attached to the last layer.
Let the extracted five-level RGB features be $(X_{i}^{r})_{i=1}^5$, depth features be $(X_{i}^{d})_{i=1}^5$, and OF features be $(X_{i}^{f})_{i=1}^5$. Their channel numbers are aligned to 64 via the compression module (CP) for computational simplicity. CP is practically realized by the $BConv$ operation mentioned later. Next, we treat RGB as the main modality, with depth and OF serving as two auxiliary modalities. Their features are handled unequally in multi-modal attention module (MAM) and refinement fusion module (RFM). The former enhances features, and the latter refines and fuses features. We also optimize the low-level features through holistic multi-modal attentive paths (HMAPs), and consistently employ a universal interaction module (UIM) in the stages of attention map interaction and progressive fusion within RFM and HMAPs. Details are described below.

\subsection{Multi-modal Attention Module (MAM)} \label{MAM}
Long-range information is well-known for enhancing features, which can be captured by the non-local (NL) module \cite{wang2018non}. Considering that NL only explores long-range information of one modality, we design MAM based on NL to model multi-modal long-range dependencies for enhancing trimodal features. Due to relatively high computation cost of the NL-like mechanism, MAM only works on the higher three hierarchies $X_{i}^{p}\in \mathbb{R}^{H \times W \times C}(p \in\{r, d, f\}, i=3,4,5)$, where $H$,$W$,$C$ refer to the height, width, and channel number, respectively. Fig.\:\ref{fig_blockdiagram} (b) shows the inner structure of MAM, where the main RGB feature are interacted with either auxiliary features $X_{i}^{m}(m \in\{d, f\})$ in the attention block. Specifically,
we first combine the main RGB features and auxiliary features into multi-modal features, as $X_{i}^{h} = BConv([X_{i}^{r},X_{i}^{m}])$, where $[\cdot,\cdot]$ denotes channel concatenation, and $BConv$ comprises convolution,  BatchNorm and ReLU. Following \cite{wang2018non}, $1 \times 1$ convolutions (Conv1 in Fig.\:\ref{fig_blockdiagram} (b)) are used to embed $X_{i}^{h}$, yielding $W_{\theta}$ and $W_{\phi}$. Then pairwise relationships of multi-modal information are captured by ``query-key matching'' as:
\setlength{\abovedisplayskip}{4pt}
\setlength{\belowdisplayskip}{4pt}
\begin{align}
    W_{a} &= softmax(W_{\theta} \otimes W_{\phi}^{T}),
\end{align}
where $\otimes$ is matrix multiplication and $W_{a}\in \mathbb{R}^{HW \times H W}$.
Further, multi-modal long-range dependencies are modeled by transmitting the multi-modal affinity matrix $W_{a}$ to the main RGB and auxiliary feature embedding, denoted as $W_{g}^{r}$ and $W_{g}^{m}$ in Fig.\:\ref{fig_blockdiagram} (b):
\begin{align}
    \hat{X}_{i}^{rm} \leftarrow W_{a}\otimes W_{g}^{r},~\hat{X}_{i}^{m} \leftarrow W_{a}\otimes W_{g}^{m},
\end{align}
where $\hat{X}_{i}^{rm}$, $\hat{X}_{i}^{m}$ are cross-modal long-range attended features (Conv1 and reshaping after $\otimes$ is omitted, and $m \in\{d, f\}$).

For either auxiliary modality, the enhanced features are obtained by adding $\hat{X}_{i}^{m}$ back to the original features, namely $Y_{i}^{m} = X_{i}^{m} + \hat{X}_{i}^{m}$.
Meanwhile, for the main RGB modality, we further concatenate the results assisted by the two auxiliary modalities, and then transform and aggregate the features using $BConv$, yielding $Y_{i}^{r}$ as shown in Fig.\:\ref{fig_blockdiagram} (b) left. In this way, all three modality-aware features are fully enhanced by propagating cross-modal long-range information.

Note in this trimodal task, we find that cross-modal long-range information is more powerful for feature enhancement than self long-range one. Meanwhile, we design RGB to be the main modality because it is more stable/reliable and serves as basic for saliency detection \cite{li2019mga, fan2020bbs}. Ablation experiments in Sec.\:\ref{Abl.3} will validate the above design ideas of MAM.

\subsection{Refinement Fusion Module (RFM)} \label{RFM}
There remain noises in each modality-aware features, which are detrimental to accurate saliency maps. Directly fusing trimodal features may lead to contamination. RFM is proposed to address this issue and is applied on all hierarchies as shown in Fig.\:\ref{fig_blockdiagram} (a). The initial RFM module in \cite{DCTNet} employed basic element-wise operations like multiplication and addition to integrate information from the three modalities, which has certain limitation in noise suppression and risks fusion-induced contamination. Additionally, the previous RFM emphasized channel attention \cite{hu2018squeeze}, overlooking valuable spatial attentive information. In this journal version, we introduce a new design of RFM to improve cross-modal interaction and fusion.

As shown in Fig.\:\ref{fig_blockdiagram} (c), we aim to achieve interactive fusion of the attention maps in RFM, as well as progressive fusion of the features from the three modalities. To tackle this, we first introduce a universal interaction module (UIM).

\begin{figure}
    \includegraphics[width=.48\textwidth]{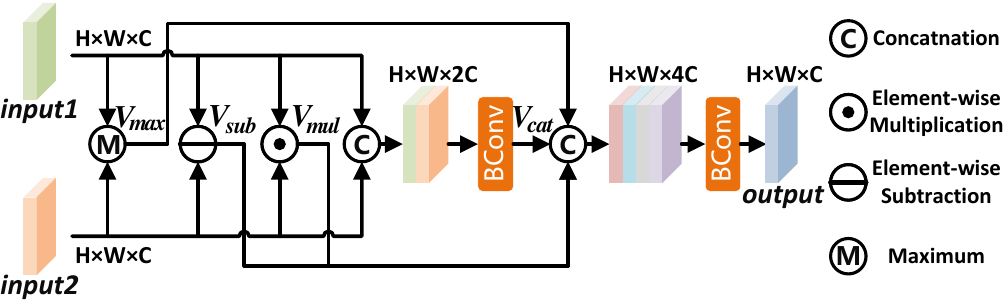}
    \vspace{-0.3cm}
    \caption{\small Universal Interaction Module (UIM) in Fig.\:\ref{fig_blockdiagram} }
    \label{fig_UIM}
    \vspace{-0.4cm}
\end{figure}

\textbf{Universal Interaction Module (UIM).} We design a universal interaction module to efficiently combine two tensors, generating a fully interacted tensor of the same size. UIM enhances attention/feature maps, incorporates complementary modality information and facilitates comprehensive integration of features. Given its plug-and-play characteristic, it holds the potential for application in various models. Specifically, UIM leverages four
standard tensor interactions as below: Concatenation is to preserve the original features and align the number of channels using the $BConv$ operation; Multiplication highlights the common features and suppresses noises; The maximum operation makes the two inputs aggregated and preserve useful feature channels; Subtraction captures the complementary information between the two tensors. Suppose their outcomes are represented by $V_{cat}$, $V_{mul}$, $V_{max}$ and $V_{sub}$, respectively. Instead of using a single operation, we amalgamate these four operations to acquire diversified feature characteristics. As depicted in Fig.\:\ref{fig_UIM}, these four outcomes are concatenated and the output channel number is enforced by a $BConv$ operation. The UIM process can be formulated as:
\begin{align}
    output &= BConv([V_{cat},V_{mul},V_{max},V_{sub}]).
\end{align}
We utilize a kernel size of 3 with both padding and stride set to 1 to keep the tensor size equivalent to the input one.

Our RFM is then realized through broad utilization of UIMs. Specifically, trimodal features are first adapted by $BConv$ for subsequent processing, yielding aligned features ${T}_{i}^{p}\ (p \in\{r, f, d\}, i=1,...,5)$.
To ensure the network emphasizes salient object information while disregarding background textures, we employ attention mechanisms to augment feature refinement. We opt for coordinate attention\cite{coa} due to its capability to extract saliency-related information for each modality across channels and also capture position-aware information. ${T}_{i}^{p}$ are augmented with the obtained coordinate attention maps of the three modalities by multiplication:
\begin{align}
    & Att_{i}^{p} = CoA(T_{i}^{p}), \\
    & \hat{T}_{i}^{p} = Att_{i}^{p} \odot {T}_{i}^{p}, ~p \in\{r, f, d\}, i=1,...,5,
\end{align}
where $CoA(\cdot)$ is the coordinate attention computation to obtain the attention map and $\odot$ denotes element-wise multiplication.

Considering the common or complementary features across diverse modalities, we employ the UIMs to acquire cross-modal coordinate attention. This involves combining the coordinate attention maps from the main RGB modality with those from the auxiliary modalities (OF and depth):
\begin{align}
    Att_{i}^{rm} &= UIM(Att_{i}^{r},Att_{i}^{m}), ~m \in\{f, d\}.
\end{align}
We employ cross-modal coordinate attention to enhance ${T}_{i}^{p}$ and add it to the results post individual modality coordinate attention augmentation. This yields richer and more accurate salient object information, promoting coordination and complementary augmentation across modalities. To maintain network stability and prevent erroneous feature suppression, residual connections are integrated to preserve original features. The above operations in RFM are summarized as:
\begin{align}
    & \hat{C}_{i}^{m} = Att_{i}^{rm} \odot {T}_{i}^{m} + \hat{T}_{i}^{m} + {T}_{i}^{m}, ~m \in\{f, d\},\\
    & C_{i}^{rm} = Att_{i}^{rm} \odot {T}_{i}^{r} + \hat{T}_{i}^{r}, ~m \in\{f, d\}, \\
    & \hat{C}_{i}^{r} = Bconv([C_{i}^{rf},C_{i}^{rd}]) + {T}_{i}^{r},
\end{align}
where ``$+$'' here denotes element-wise addition. As RGB serves as the main modality, we acquire two features enhanced through cross-modal attention. We concatenate and aggregate them via convolution, and apply a residual connection.


Coordinate attention primarily captures cross-channel and coarse spatial salient information. Subsequently, spatial attention\cite{woo2018cbam} is employed to encode local spatial information, concentrating on salient areas for local refinement. The operations of spatial attention are similar to coordinate attention, as shown in Fig.\:\ref{fig_blockdiagram} (c). Hence we do not reiterate the explanation.


Let the features of the three modalities after refinement be ${Z}_{i}^{r}$, ${Z}_{i}^{f}$ and ${Z}_{i}^{d}$. They are fused progressively again using UIM to achieve the merged features:
\begin{align}
    F_{i} &= UIM({Z}_{i}^{r},UIM({Z}_{i}^{f},{Z}_{i}^{d})).
\end{align}
Progressive fusion facilitated by UIM establishes advantageous in extracting valuable information from two auxiliary modalities and subsequently achieving extensive interaction with the RGB modality. Initially, we merge their features using UIM and then combine the outcome with RGB features. In the ablation experiments (Sec.\:\ref{Abl.4}), it outperforms the equivalent fusion of three features. This approach extends beyond the RFM module and is also employed in generating a coarse map for HMAPs (detailed later).

Finally, hierarchical fused features are ×2 bilinear interpolated to match the lower-level layer (optionally), and they are concatenated and processed by $BConv$, composing a typical U-Net\cite{2015unet} decoder to achieve final saliency prediction.

\subsection{Holistic Multi-modal Attentive Paths (HMAPs)} \label{Refine}
As illustrated in Fig.\:\ref{fig_blockdiagram} (a), our approach, inspired by \cite{f3net}, aims to improve the precision of the input features preceding the RFM module through the attenuation of undesired noise. It is widely acknowledged that the initial three encoder layers preserve substantial low-level information pivotal for fine-grained detection, albeit containing lots of noises.

To address this, HMAPs are deliberately positioned ahead of the RFM modules. This strategic placement is motivated by the phenomenon observed as the network increases in depth, where the information content gradually wanes and the corresponding refinement effect becomes less pronounced. This is empirically supported by our ablation experiments (Sec.\ref{Abl.5}). Furthermore, to achieve this objective, we employ the \emph{LastConv}, aimed at consolidating the high-level features following the ASPP into a single channel, yielding $L^{r}$, $L^{f}$ and $L^{d}$ in Fig.\:\ref{fig_blockdiagram} (a). \emph{LastConv} comprises two \emph{BConv} modules and a convolutional layer, ultimately generating a single output channel while retaining the original feature size intact. Then we progressively fuse them to predict a coarse map:
\begin{align}
    CM &= UIM(L^{r},UIM(L^{f},L^{d})),
\end{align}
where $CM$ denotes the coarse map.
Subsequently, we employ the coarse map for enhancing low-level features while concurrently mitigating noise interference:
\begin{align}\label{equ_hmaps}
    f_{i}^{p} &= f_{i}^{p} \odot CM + f_{i}^{p},
\end{align}
where $f_{i}^{p} \in \{X_{1}^{p},X_{2}^{p},Y_{3}^{p}\}$, $~p \in\{r, f, d\}$, and $\odot$ represents broadcast multiplication.
Equation (\ref{equ_hmaps}) elucidates the Interaction Enhancement (red solid circle) depicted in Fig.\:\ref{fig_blockdiagram}, which symbolizes the refinement of low-level features by the coarse map through element-wise multiplication and addition.

The motivation for refining the initial three layers is to minimize the introduction of noise from the coarse map into high-level features. Ablation experiments (Sec.\:\ref{Abl.5}) validate the advantages of refining these first three layers, as opposed to all five. While processing the first two layers, we uphold a consistent feature size through ×2 bilinear interpolation.

\subsection{Supervision}  \label{Supervision}
We adopt a combination of widely used binary cross entropy (BCE) loss and intersection-over-union (IoU) loss \cite{iouloss} for training DCTNet+. The total loss is formulated as: $l_{total} = \sum_{i = 1}^{5} ({1}/{2^{i-1}}) l(S_{i},G) + l(CM,G)$, where $l(S_{i},G) = l_{bce}(S_{i},G) + l_{iou}(S_{i},G)$.
$S_{i}$ signifies the output (resized back to the input size) derived from the $i$-th layer of the decoder, $G$ represents the ground truth, and $l_{bce}(S_{i},G)$ and $l_{iou}(S_{i},G)$ stand for the BCE loss and IoU loss, respectively. We incorporate supervision into the coarse map to guide it for improved performance. Note we assign higher-level loss the lower weight (\emph{i.e.}, ${1}/{2^{i-1}}$) due to its larger error. During inference, we take $S_{1}$ as the final saliency prediction.

\section{EXPERIMENTS}
\label{sec:Experiments}

\begin{table*}[t!]
    \renewcommand{\arraystretch}{1.0}
    \caption{\small Quantitative comparison with state-of-the-art VSOD methods on 5 benchmark datasets. The top three results are represented in \textcolor{red}{red}, \textcolor{green}{green}, and \textcolor{blue}{blue} from top to bottom, respectively. $\uparrow$/$\downarrow$ denotes that the larger/smaller value is better. Symbol `**' means that results are not available.
    }\label{tab_sota}
    \vspace{-8pt}
  \centering
    \footnotesize
    \setlength{\tabcolsep}{1.2mm}

    \begin{tabular}{c|c|ccc|ccc|ccc|ccc|ccc}
    \hline
    \multirow{2}{*}{Methods}&\multirow{2}{*}{Year}&\multicolumn{3}{c|}{DAVIS\cite{davis}}&
    \multicolumn{3}{c|}{DAVSOD\cite{fan2019shifting}}&\multicolumn{3}{c|}{FBMS\cite{fbms}}&\multicolumn{3}{c|}{SegV2\cite{segv2}}&\multicolumn{3}{c}{VOS\cite{vos}}\cr\cline{3-17}
    &&$F_{\beta}^{\textrm{max}}\uparrow$&$S_\alpha\uparrow$&$M\downarrow$&$F_{\beta}^{\textrm{max}}\uparrow$&$S_\alpha\uparrow$&$M\downarrow$&$F_{\beta}^{\textrm{max}}\uparrow$&$S_\alpha\uparrow$&$M\downarrow$&$F_{\beta}^{\textrm{max}}\uparrow$&$S_\alpha\uparrow$&$M\downarrow$&$F_{\beta}^{\textrm{max}}\uparrow$&$S_\alpha\uparrow$&$M\downarrow$\cr
    \hline

    MSTM\cite{mstm}&CVPR'16&0.395&0.566&0.174&0.347&0.530&0.214&0.500&0.613&0.177&0.526&0.643&0.114&0.567&0.657&0.144 \cr
    STBP\cite{xi2016stbp}&TIP'16&0.485&0.651&0.105&0.408&0.563&0.165&0.595&0.627&0.152&0.640&0.735&0.061&0.526&0.576&0.163 \cr
    SFLR\cite{chen2017sflr}&TIP'17&0.698&0.771&0.060&0.482&0.622&0.136&0.660&0.699&0.117&0.745&0.804&0.037&0.546&0.624&0.145\cr
    SCOM\cite{scom}&TIP'18&0.746&0.814&0.055&0.473&0.603&0.219&0.797&0.794&0.079&0.764&0.815&0.030&0.690&0.712&0.162 \cr
    SCNN\cite{scnn}&TCSVT'18&0.679&0.761&0.077&0.494&0.680&0.127&0.762&0.794&0.095&**&**&**&0.609&0.704&0.109 \cr
    FGRNE\cite{li2018fgrne}&CVPR'18&0.783&0.838&0.043&0.589&0.701&0.095&0.767&0.809&0.088&0.694&0.770&0.035&0.669&0.715&0.097 \cr
    PDBM\cite{song2018pdbm}&ECCV'18&0.855&0.882&0.028&0.591&0.706&0.114&0.821&0.851&0.064&0.808&0.864&0.024&0.742&0.818&0.078 \cr
    SSAV\cite{fan2019shifting}&CVPR'19&0.861&0.893&0.028&0.659&0.755&0.084&0.865&0.879&0.040&0.798&0.851&0.023&0.704&0.786&0.091 \cr
    MGAN\cite{li2019mga}&ICCV'19&0.893&0.913&0.022&0.663&0.757&0.079&0.909&\textcolor{green}{0.912}&\textcolor{blue}{0.026}&0.840&\textcolor{blue}{0.895}&0.024&0.743&0.807&0.069 \cr
    PCSA\cite{gu2020pcsa}&AAAI'20&0.880&0.902&0.022&0.656&0.741&0.086&0.837&0.868&0.040&0.811&0.866&0.024&0.747&0.828&0.065 \cr
    TENet\cite{ren2020tenet}&ECCV'20&0.881&0.905&0.017&0.697&0.779&0.070&\textcolor{green}{0.915}&\textcolor{red}{0.916}&\textcolor{red}{0.024}&**&**&**&**&**&** \cr
    FSNet\cite{ji2021fsnet}&ICCV'21&0.907&0.920&0.020&0.685&0.773&0.072&0.888&0.890&0.041&0.806&0.870&0.025&0.659&0.703&0.108 \cr
    DCFNet\cite{zhang2021dcfnet}&ICCV'21&0.900&0.914&0.016&0.660&0.741&0.074&0.840&0.873&0.039&0.839&0.883&\textcolor{blue}{0.015}&\textcolor{blue}{0.791}&\textcolor{blue}{0.845}&\textcolor{blue}{0.061} \cr
    UGPL\cite{piao2022ugpl}&NeurIPS'22&0.895&0.910&0.020&0.658&0.749&0.074&0.892&0.900&0.027&0.803&0.860&0.025&0.706&0.764&0.078 \cr
    MGTNet\cite{min2022mgtnet}&SPL'22&\textcolor{green}{0.918}&\textcolor{green}{0.925}&\textcolor{blue}{0.015}&0.721&0.796&\textcolor{blue}{0.064}&0.890&0.901&0.033&\textcolor{blue}{0.849}&0.893&\textcolor{green}{0.014}&0.766&0.835&0.062 \cr
    DMPNet\cite{chen2023dmpnet}&TOMM'23&0.888&0.905&0.021&0.655&0.746&0.069&0.888&0.894&0.038&**&**&**&**&**&** \cr
    CoSTFormer\cite{liu2023costformer}&TNNLS'23&0.903&0.921&\textcolor{green}{0.014}&\textcolor{green}{0.731}&\textcolor{green}{0.806}&\textcolor{green}{0.061}&0.885&0.889&0.036&\textcolor{green}{0.870}&\textcolor{green}{0.904}&0.016&0.748&0.812&0.081 \cr
    DCTNet\cite{DCTNet}&ICIP'22&\textcolor{blue}{0.912}&\textcolor{blue}{0.922}&\textcolor{blue}{0.015}&\textcolor{blue}{0.728}&\textcolor{blue}{0.797}&\textcolor{green}{0.061}&\textcolor{blue}{0.913}&\textcolor{blue}{0.911}&\textcolor{green}{0.025}&0.840&0.889&0.019&\textcolor{green}{0.793}&\textcolor{green}{0.846}&\textcolor{red}{0.051} \cr
    DCTNet+& -- &\textcolor{red}{0.922}&\textcolor{red}{0.930}&\textcolor{red}{0.012}&\textcolor{red}{0.754}&\textcolor{red}{0.818}&\textcolor{red}{0.055}&\textcolor{red}{0.918}&\textcolor{red}{0.916}&\textcolor{blue}{0.026}&\textcolor{red}{0.917}&\textcolor{red}{0.931}&\textcolor{red}{0.010}&\textcolor{red}{0.802}&\textcolor{red}{0.858}&\textcolor{green}{0.056} \cr
    \hline
    \end{tabular}
\vspace{-0.4cm}
\end{table*}

\subsection{Datasets and Metrics}\label{sec51}

Although the proposed RDVS dataset provides realistic depth of the videos, there exist various VSOD benchmark datasets as well as models. To enable comparison with those models, we consider generating synthetic depth \cite{DCTNet} for these common RGB videos and convert them into ``pseudo'' RGB-D videos, in order for more comprehensive evaluation and comparisons of the proposed scheme. To obtain a synthetic depth map for each video frame, DPT \cite{dpt} is adopted according to our previous work \cite{DCTNet}.
We conduct experiments on five VSOD benchmark datasets, namely DAVIS \cite{davis}, DAVSOD (easy-35 test set) \cite{fan2019shifting}, FBMS \cite{fbms}, SegTrack-V2 \cite{segv2}, and VOS \cite{vos}, as well as on the proposed RDVS dataset for RGB-D VSOD. When evaluating DCTNet+ on those VSOD benchmarks, we select the training sets of DAVIS, DAVSOD, FBMS, and DUTS \cite{duts} for model training, with their depth maps being synthesized. Note that DUTS is a still image dataset for salient object detection, and is universally adopted to pre-train VSOD models \cite{li2019mga, ji2021fsnet}. The training sets of DAVIS, DAVSOD and FBMS contain 30, 61 and 29 sequences, respectively. For evaluation on RDVS, two protocols are adopted. The first one is to evaluate on the entire RDVS, while the second one is to first conduct fine-tuning on the training set of RDVS, and then test on RDVS's test set. More details refer to Sec.\:\ref{RDVSExperiments}.

For performance evaluation, we employ three widely used evaluation metrics, namely S-measure ($S_\alpha$) \cite{smeasure},  maximum F-measure ($F_{\beta}^{\textrm{max}}$) \cite{borji2015benchmark,maxf}, and MAE ($M$) \cite{borji2015benchmark,mae}.
Higher $S_\alpha$, $F_{\beta}^{\textrm{max}}$, and lower $M$ indicate better performance.

\subsection{Implementation Details}\label{sec52}
Our DCTNet+ is implemented by PyTorch. Following a similar training strategy as \cite{li2019mga,ji2021fsnet}, we first pre-train the depth stream on synthetic depth maps and the OF stream on rendered OF images. We also employ RGB images to pre-train the RGB stream. Finally the entire model is fine-tuned on the training set. Trimodal inputs of RGB, depth and OF are resized to $448 \times 448 $ with batch size 8. We adopt the SGD algorithm to optimize. The initial learning rates of backbones and other parts are set to 1e-4 and 1e-3, respectively. Data augmentation like random flipping, random cropping are utilized. We train on an NVIDIA 3090 GPU for a total of 30 epochs, taking $\sim$17 hours.

\begin{figure*}
    \includegraphics[width=1.0\textwidth]{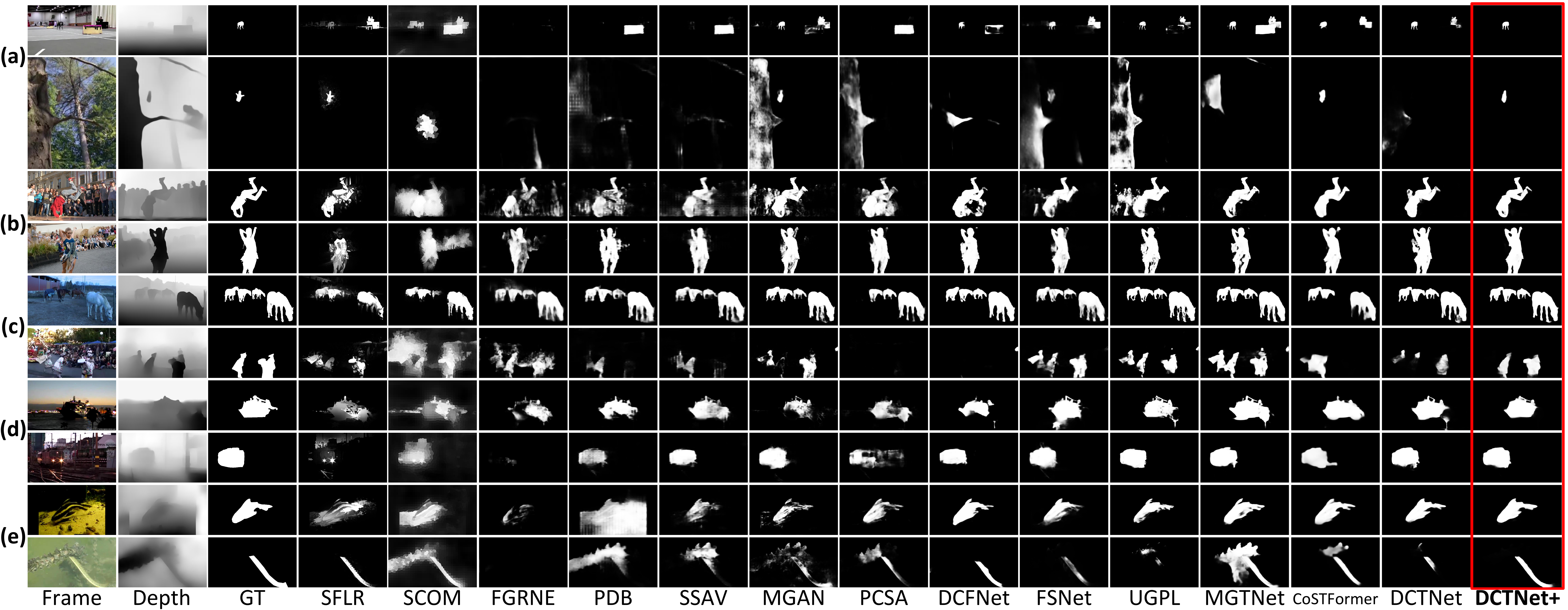}\vspace{-0.3cm}
    \caption{\small Qualitative comparison of our DCTNet+ model and SOTA methods on conventional VSOD benchmarks.}
    \label{visual}\vspace{-0.5cm}
\end{figure*}

\subsection{Comparison to State-of-the-Arts on VSOD Benchmarks}\label{sec53}

To validate the proposed DCTNet+, we first compare it on conventional VSOD benchmarks with 17 state-of-the-art (SOTA) methods, including 4 traditional methods \cite{mstm,xi2016stbp,chen2017sflr,scom}, and 13 deep learning methods \cite{scnn,li2018fgrne,song2018pdbm,fan2019shifting,li2019mga,gu2020pcsa,ren2020tenet,ji2021fsnet,zhang2021dcfnet,piao2022ugpl,min2022mgtnet,chen2023dmpnet,liu2023costformer}. In particular, notation ``DCTNet'' denotes our previous model in \cite{DCTNet}, and ``DCTNet+'' denotes the improved model presented in this paper. Quantitative results on five widely used datasets are given in Table\:\ref{tab_sota}. DCTNet+ achieves the best results on both $F_{\beta}^{\textrm{max}}$ and $S_{\alpha}$.
Its MAE metrics are also impressive for DAVIS, DAVSOD, and SegTrack-V2, and achieve the top three results on the other two datasets.
When compared with DCTNet, DCTNet+ achieves substantially notable improvements on all datasets and on almost all metrics, showing that our new designs of RFM, HMAPs and additional supervision are favorable.

For qualitative evaluation, in Fig.\:\ref{visual} we show a visual comparison between DCTNet+, DCTNet and other SOTA contenders. It can be seen that DCTNet+ generates more accurate saliency images with background being well suppressed. Specifically, Fig.\:\ref{visual} (a)'s cases are fast-moving small objects, while all other SOTA methods mis-recognize background clutters as salient objects. Only DCTNet+ correctly identifies them; Likewise, Fig.\:\ref{visual} (b)-(e) include cases of complex background, multiple objects, low illumination and low contrast, respectively, from which one can see that DCTNet+ achieves more competitive results.

\subsection{Comparison on the Proposed RDVS Dataset}
\label{RDVSExperiments}

\subsubsection{Straightforward evaluation on the full RDVS dataset}\label{rdvs.1}

\begin{table}[t!]

  \renewcommand{\arraystretch}{1.0}
  \vspace{0.1cm}
    \caption{Results of SOTA methods in different fields and the proposed method on RDVS dataset, where the suffix "$\divideontimes$" indicates RGB-D VSOD techniques, "$\uparrow/\downarrow$" denotes that the larger/smaller value is better. The best are stressed in \bf{BOLD}.}\label{tab_rdvs1}
    \vspace{-8pt}
   \centering
    \footnotesize
    \setlength{\tabcolsep}{2.35mm}
    \begin{tabular}{c|c|c|ccc}
    \hline
    &Methods&Year&$F_{\beta}^{\textrm{max}}\uparrow$&$S_\alpha\uparrow$&$M\downarrow$\cr
    \hline
    \multirow{11}{*}{\begin{sideways}{RGB-D SOD}\end{sideways}}
    &BBSNet\cite{fan2020bbs}&ECCV'20&0.679&0.790&0.054\cr
    &JL-DCF\cite{fu2020jl-dcf}&CVPR'20&0.691&0.789&0.058\cr
    &S2MA\cite{liu2020s2ma}&CVPR'20&0.638&0.758&0.076\cr
    &HDFNet\cite{pang2020hdfnet}&ECCV'20&0.713&0.800&0.056\cr
    &DPANet\cite{chen2020dpanet}&TIP'20&0.714&0.803&0.055\cr
    &BTSNet\cite{zhang2021bts}&ICME'21&0.699&0.795&0.057\cr
    &SPNet\cite{zhou2021specificity}&ICCV'21&0.699&0.793&0.055\cr
    &TriTransNet\cite{liu2021tritransnet}&ACMMM'21&0.697&0.793&0.051\cr
    &CDNet\cite{jin2021cdnet}&TIP'21&0.691&0.795&0.057\cr
    &DCF\cite{ji2021calibrated}&CVPR'21&0.679&0.778&0.060\cr
    &RD3D+\cite{chen20223d}&TNNLS'22&0.725&0.814&0.052\cr
    &CIRNet\cite{CIRNet}&TIP'22&0.693&0.785&0.062\cr
    &PICRNet\cite{PICR-Net}&ACMMM'23&0.693&0.787&0.059\cr
    &HRTransNet\cite{Tang2023RTransNet}&TCSVT'23&0.700&0.794&0.050\cr
    \hline
    \multirow{6}{*}{\begin{sideways}{VSOD}\end{sideways}}
    &PDB\cite{song2018pdbm}&ECCV'18&0.704&0.797&0.059\cr
    &MGAN\cite{li2019mga}&ICCV'19&0.796&0.847&0.048\cr
    &SSAV\cite{fan2019shifting}&CVPR'19&0.737&0.814&0.056\cr
    &PCSA\cite{gu2020pcsa}&AAAI'20&0.545&0.705&0.082\cr
    &FSNet\cite{ji2021fsnet}&ICCV'21&0.773&0.829&0.053\cr
    &DCFNet\cite{zhang2021dcfnet}&ICCV'21&0.742&0.811&0.050\cr
    \hline
    \multirow{2}{*}{$\divideontimes$}
    &DCTNet\cite{DCTNet}&ICIP'22&0.816&0.854&0.044\cr
    &DCTNet+&--& \bf 0.836& \bf 0.871& \bf 0.042\cr
    \hline
    \end{tabular}
    \vspace{-0.4cm}
\end{table}

\begin{figure}
    \includegraphics[width=.48\textwidth]{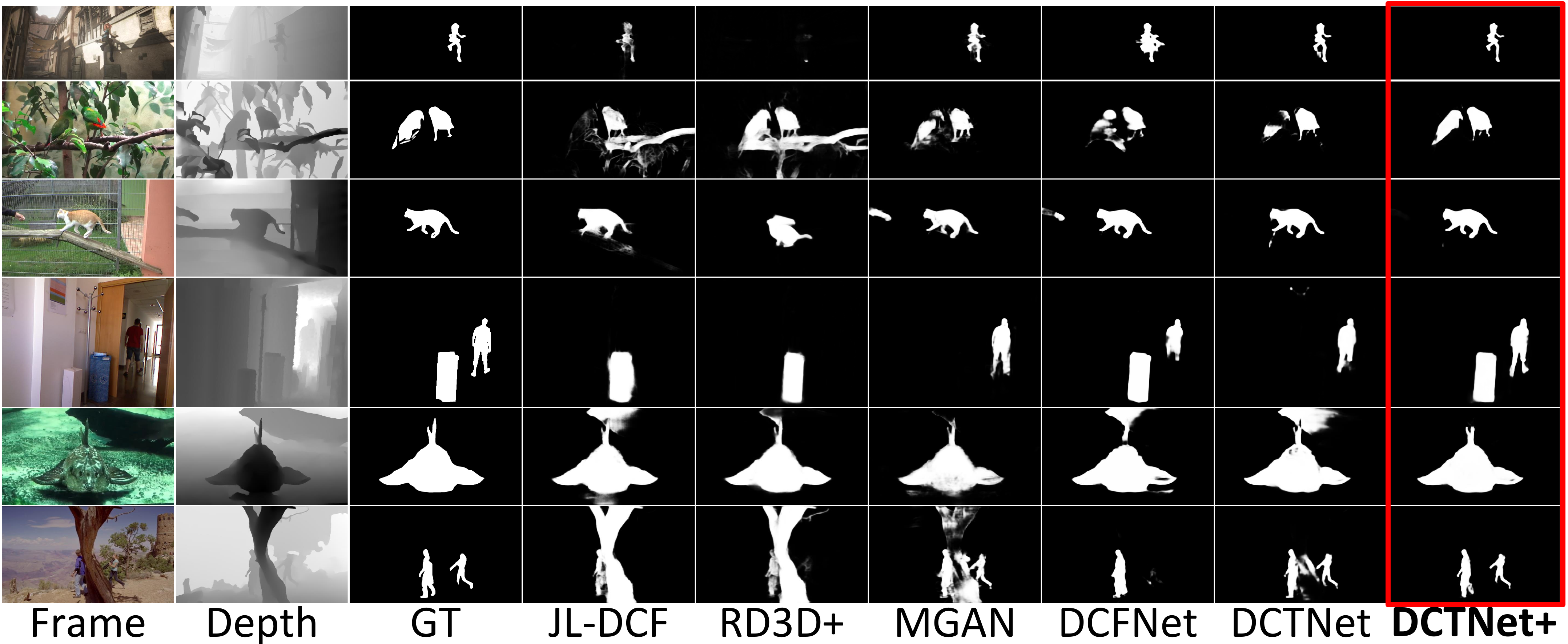}
    \vspace{-0.3cm}
    \caption{\small Qualitative comparison on the proposed RDVS dataset.}
    \label{RDVSVisual}
    \vspace{-0.4cm}
\end{figure}

As aforementioned in Sec.\:\ref{dataset splits}, this protocol sets the entire RDVS dataset for testing, which means directly testing existing already-trained models on the whole RDVS without fine-tuning. Such a treatment is similar to that of many benchmark datasets (like SIP or SegV2) in RGB-D/video SOD literature, where the datasets are just employed for testing. Since RDVS's depth is realistic, it enables us to test some RGB-D SOD models as well.
To this end, we opted for 11 publicly available RGB-D SOD methods\cite{fan2020bbs,chen20223d,fu2020jl-dcf,zhang2021bts,liu2020s2ma,pang2020hdfnet,chen2020dpanet,zhou2021specificity,liu2021tritransnet,jin2021cdnet,ji2021calibrated,PICR-Net,Tang2023RTransNet,CIRNet}, and 6 publicly available VSOD methods\cite{song2018pdbm,li2019mga,fan2019shifting,gu2020pcsa,ji2021fsnet,zhang2021dcfnet}, together with DCTNet\cite{DCTNet} and the proposed DCTNet+. The experimental results and visual comparison results are shown in Table\:\ref{tab_rdvs1} and Fig.\:\ref{RDVSVisual}.

Among RGB-D SOD models, we note that RD3D+\cite{chen20223d} exhibits the best overall performance on RDVS, and this may be because that RD3D+ employs 3D convolutions, enabling pre-fusion of modalities at the encoder stage and deep fusion at the decoder stage, effectively promoting fusion between RGB and depth information. In addition, both DPANet \cite{chen2020dpanet} and S2MA \cite{liu2020s2ma} utilizing non-local mechanisms perform decently on RDVS, whereas the former achieves slightly better accuracy attributed to the consideration of depth quality.

For VSOD models, they demonstrate higher performance on RDVS compared to those RGB-D SOD models, emphasizing that the RGB-D VSOD task relies still on video data and necessitates more support from motion information. Among the VSOD models, MGAN \cite{li2019mga} exhibits the most superior accuracy, whereas DCFNet \cite{zhang2021dcfnet} proposed later lags behind MGAN.
This is probably caused by the single-stream architecture of DCFNet, while MGAN adopts a dual-stream network leveraging optical flow.
This phenomenon reflects that explicitly incorporating optical flow as temporal cues could somewhat guarantee VSOD accuracy and is a prevailing technical scheme, as also designed in FSNet \cite{ji2021fsnet} as well as DCTNet+. PCSA \cite{gu2020pcsa} is the least effective due to its lightweight nature and therefore less generalization.

In contrast, the results from DCTNet and DCTNet+ showcase top performance on RDVS, surpassing all existing approaches from both domains. This reveals a crucial issue for the RGB-D VSOD task: promising results can be achieved by harnessing temporal, spatial, together with depth information. Therefore, future related research may focus on how to utilize such trimodal cues to pursue more powerful RGB-D VSOD.

\begin{table}[t!]
  \renewcommand{\arraystretch}{1.0}
  \vspace{0.1cm}
    \caption{Results of SOTA methods in different fields as well as the proposed method on RDVS testing set. The left half are the results of the original model applied directly on the RDVS testing set, and the right half are the results obtained by re-training the models consistently on the RDVS training set and then evaluating them on the RDVS testing set. The best are highlighted in \bf{BOLD}.}\label{tab_rdvs2}
    \vspace{-8pt}
   \centering
    \footnotesize
    \setlength{\tabcolsep}{1.40mm}
    \begin{tabular}{c|c|ccc|ccc}
    \hline
    &\multirow{2}{*}{Methods}&\multicolumn{3}{c|}{Original model}&\multicolumn{3}{c}{Fine-tuned model}\cr\cline{3-8}    &&$F_{\beta}^{\textrm{max}}\uparrow$&$S_\alpha\uparrow$&$M\downarrow$&$F_{\beta}^{\textrm{max}}\uparrow$&$S_\alpha\uparrow$&$M\downarrow$\cr
    \hline
    \multirow{6}{*}{\begin{sideways}{RGB-D SOD}\end{sideways}}
    &BBSNet\cite{fan2020bbs}&0.549&0.732&0.056&0.642&0.778&0.049\cr
    &RD3D+\cite{chen20223d}&0.607&0.764&0.056&0.681&0.799&0.054\cr
    &JL-DCF\cite{fu2020jl-dcf}&0.559&0.725&0.067&0.570&0.734&0.068\cr
    &BTSNet\cite{zhang2021bts}&0.573&0.738&0.061&0.636&0.769&0.060\cr
    &SPNet\cite{zhou2021specificity}&0.570&0.736&0.063&0.654&0.782&0.048\cr
    &TriTransNet\cite{liu2021tritransnet}&0.559&0.728&0.060&0.586&0.732&0.053\cr
    \hline
    \multirow{3}{*}{\begin{sideways}{VSOD}\end{sideways}}
    &MGAN\cite{li2019mga}&0.736&0.826&0.043&0.720&0.822&0.040\cr
    &FSNet\cite{ji2021fsnet}&0.745&0.824&0.046&0.697&0.800&0.047\cr
    &DCFNet\cite{zhang2021dcfnet}&0.647&0.768&0.049&0.650&0.774&0.054\cr
    \hline
    \multirow{2}{*}{$\divideontimes$}
    &DCTNet\cite{DCTNet}&0.780&0.846& \bf 0.033&0.787&0.849&0.035\cr
    &DCTNet+& \bf 0.803& \bf 0.861&0.036& \bf 0.814& \bf 0.869& \bf 0.034\cr
    \hline
    \end{tabular}
    \vspace{-0.4cm}
\end{table}

\subsubsection{Evaluation on RDVS test set after fine-tuning}\label{rdvs.2}
Considering that the aforementioned evaluations in Table\:\ref{tab_rdvs1} are for models pre-trained on specific datasets within their domains, the comparison may have some limitations. To address this and using the second protocol, we have fine-tuned the models on the defined RDVS training dataset. The fine-tuning process involves loading pre-trained weights of different models, ensuring that they have good initialization for saliency detection, and then fine-tune them on the RDVS training set consisting of 2,208 frames to align with the data distribution of RDVS. Subsequently, the obtained results on RDVS test set are shown in Table\:\ref{tab_rdvs2}. Note for convenience of comparison in Table\:\ref{tab_rdvs2}, we also show the results by directly applying the original models prior to fine-tuning (\emph{i.e.}, the corresponding models exhibited in Table\:\ref{tab_rdvs1}) on this test set.

From Table\:\ref{tab_rdvs2}, one can see that methods after fine-tuning generally show better performance compared to those without fine-tuning, particularly for the cases of RGB-D SOD models. This issue could happen due to the discrepancy in depth sources, namely the sources of depth maps used for RGB-D SOD, and the depth sources in RDVS. Note that RDVS encompasses diverse depth sources including those from computer graphics, structured light, and binocular computation, contributing to distinct depth data domains from those previously adopted in RGB-D SOD.
Fine-tuning RGB-D SOD models hence facilitates domain adaptation on RDVS, and yields more precise SOD.
Meanwhile, for those VSOD models, there exists no explicit domain discrepancy as they do not take depth inputs, fine-tuning them in turn offers limited improvement and can even lead to over-fitting and downgrade the performance, \emph{e.g.} the cases of MGAN and FSNet.

In summary, the comparative analyses suggest that while the VSOD models generally outperform the RGB-D SOD models, the gaps between them diminish. This phenomenon further demonstrates the benefit of depth in the RGB-D VSOD context. Moreover, our DCTNet+ continues to yield the most favorable results, showing the power of incorporating comprehensive and complementary cues for RGB-D VSOD.

\subsubsection{Synthetic depth \emph{v.s.} realistic depth}\label{rdvs.3}

\begin{table}[t!]

  \renewcommand{\arraystretch}{1.0}
  \vspace{0.1cm}
    \caption{Experimental results of comparing synthetic depth maps and realistic depth maps by applying the original models to the full RDVS dataset. The best are highlighted in \bf{BOLD}.}\label{tab_rdvs3}
    \vspace{-8pt}
   \centering
    \footnotesize
    \setlength{\tabcolsep}{1.80mm}
    \begin{tabular}{c|ccc|ccc}
    \hline
    \multirow{2}{*}{Methods}&\multicolumn{3}{c|}{Synthetic depth}&\multicolumn{3}{c}{Realistic depth}\cr\cline{2-7}    &$F_{\beta}^{\textrm{max}}\uparrow$&$S_\alpha\uparrow$&$M\downarrow$&$F_{\beta}^{\textrm{max}}\uparrow$&$S_\alpha\uparrow$&$M\downarrow$\cr
    \hline
    BBSNet\cite{fan2020bbs}&0.690&0.794&0.056&0.679&0.790&0.054\cr
    JL-DCF\cite{fu2020jl-dcf}&0.707&0.804&0.055&0.691&0.789&0.058\cr
    BTSNet\cite{zhang2021bts}&0.703&0.803&0.053&0.699&0.795&0.057\cr
    \hline
    $\divideontimes$DCTNet\cite{DCTNet}&0.810&0.852&0.046&0.816&0.854&0.044\cr
    $\divideontimes$DCTNet+& \bf 0.833& \bf 0.868& \bf 0.043& \bf 0.836& \bf 0.871& \bf 0.042\cr
    \hline
    \end{tabular}
    \vspace{-0.4cm}
\end{table}

\begin{table}[t!]

  \renewcommand{\arraystretch}{1.0}
  \vspace{0.1cm}
    \caption{Experimental results of comparing synthetic depth maps and realistic depth maps by fine-tuning the models on the RDVS training set. The best are highlighted in \bf{BOLD}.}\label{tab_rdvs4}
    \vspace{-8pt}
   \centering
    \footnotesize
    \setlength{\tabcolsep}{1.80mm}
    \begin{tabular}{c|ccc|ccc}
    \hline
    \multirow{2}{*}{Methods}&\multicolumn{3}{c|}{Synthetic depth}&\multicolumn{3}{c}{Realistic depth}\cr\cline{2-7}    &$F_{\beta}^{\textrm{max}}\uparrow$&$S_\alpha\uparrow$&$M\downarrow$&$F_{\beta}^{\textrm{max}}\uparrow$&$S_\alpha\uparrow$&$M\downarrow$\cr
    \hline
    BBSNet\cite{fan2020bbs}&0.603&0.761&0.057&0.642&0.778&0.049\cr
    JL-DCF\cite{fu2020jl-dcf}&0.543&0.714&0.075&0.570&0.734&0.068\cr
    BTSNet\cite{zhang2021bts}&0.613&0.764&0.059&0.636&0.769&0.060\cr
    \hline
    $\divideontimes$DCTNet\cite{DCTNet}&0.763&0.835&0.040&0.787&0.849&0.035\cr
    $\divideontimes$DCTNet+& \bf 0.796& \bf 0.857& \bf 0.039& \bf 0.814& \bf 0.869& \bf 0.034\cr
    \hline
    \end{tabular}
    \vspace{-0.4cm}
\end{table}

Since our RDVS dataset provides realistic depth for videos with SOD annotations, it enables us to compare the impact of realistic depth against the synthetic one and give some insight into the question: \emph{``Can the synthetic depth replace the realistic depth in practice?''} To this end, we consider generating synthetic depth across the entire RDVS using DPT\cite{dpt}. By feeding both synthetic and realistic depth maps, we re-evaluate only selected RGB-D SOD models together with DCTNet and DCTNet+, as those VSOD models will not be influenced.

First, following Sec.\:\ref{rdvs.1}, we perform straightforward evaluation on the full RDVS dataset, where the realistic depth is replaced with synthetic depth, and the experimental results are presented in Table\:\ref{tab_rdvs3}.
Notably, only DCTNet and DCTNet+ yield improvements when  realistic depth was fed, contrasting with the diminished performance observed for those RGB-D SOD models. Note that only this fact cannot negate the efficacy of realistic depth, since there exists domain discrepancy regarding the depth sources as discussed above. 

Next, following Sec.\:\ref{rdvs.2}, we evaluate on RDVS test set after fine-tuning, and meanwhile replace realistic depth with synthetic depth. Note that for this experiment setup, the types of depth for training and testing set are kept consistent, namely either synthetic or realistic, and the resulting evaluation metrics are shown in Table\:\ref{tab_rdvs4}.
One can see that under this protocol, all models exhibit better accuracy when realistic depth is used, and using synthetic depth for training and testing in turn degrades the performance. We attribute this to the fact that synthetic depth is generated from RGB images by algorithms and also heavily resorts to prior knowledge. This makes the synthetic depth sometimes lack genuine and complementary geometry information that is able to well compensate the RGB modality. Finally, this again highlights the value and significance of constructing a RGB-D VSOD dataset based on realistic scene depth.

\subsection{Ablation Study}\label{Ablation}
We conduct thorough ablation studies by removing or replacing components from the full implementation of DCTNet+ on DAVIS, DAVSOD, VOS, and RDVS to evaluate the contribution of each key component. Note that ablating on RDVS is essential in order to validate that the components are effective on practical RGB-D videos, and we use the aforementioned fine-tuning setup, namely testing on RDVS test set after fine-tuning on the RDVS training set.

\subsubsection{Effectiveness of incorporating depth} \label{Abl.1}
One main contribution of our work is introducing depth information into RGB videos for better SOD. To validate its rationality, we change the two proposed modules (MAM and RFM) by removing depth-related parts, and also exclude the depth stream. This retained two-stream variant is denoted as A1, which is re-trained and tested using RGB and OF. The results on four datasets are shown in Table\:\ref{tab_ab1}, which shows incorporating depth does boost performance by notable margins. Especially, the highest metric margins ($\sim$4.9\% on $F_{\beta}$) are observed on RDVS, validating the necessity of realistic depth.

\subsubsection{Choice of the main modality} \label{Abl.2}
Since we treat different modalities unequally, and take RGB as the main modality, we conduct experiments to validate this choice by choosing depth/OF as the main one, yielding variants B1 and B2. As shown in Table\:\ref{tab_ab1}, both of them achieve inferior results compared to our full DCTNet+ model.

\subsubsection{Effectiveness of MAM module} \label{Abl.3}
In order to verify the effectiveness of the proposed MAM module, we first directly remove the MAM module, denoted as variant C1, and then replace the MAM with three original non-local modules, denoted as variant C2. As shown in Table \:\ref{tab_ab1}, by comparing C1, C2 with our full model, it can be seen that our MAM delivers a higher lift due to the multi-modal long-range dependencies.

\subsubsection{Effectiveness of RFM module} \label{Abl.4}
In order to validate RFM and its inner components, as shown in Table\:\ref{tab_ab1}, we design five variants. Firstly, variant D1 replaces the entire RFM module with simple concatenation without refinement. We then replace RFM proposed in this paper with the previous RFM module in \cite{DCTNet}, yielding variant D2. Variant D3 replaces coordinate attention in RFM with the common channel attention, showing the efficacy of coordinate attention in RFM. To validate the design of coordinate attention followed by spatial attention in RFM, we switch the two and obtain variant D4. Finally, we replace progressive fusion with equal fusion and obtain variant D5, in order to validate the progressive fusion. Specifically, the equal fusion is achieved by extending UIM to take three inputs, so that the three modalities can interact and fuse simultaneously. When dealing with subtraction, a two-by-two interaction is adopted, followed by concatenation for fusion.

UIM is a universal module for feature aggregation, and we design five variants to demonstrate its inner designs. As UIM is composed of concatenation, multiplication, maximization, and subtraction, we first replace UIM with each of these four basic operations and obtain variants E1$\sim$E4, respectively. As shown in Table\:\ref{tab_ab1}, one can see that UIM is more powerful when using comprehensive basic operations. Besides, we compare UIM with the combination of multiplication and addition aggregation (E5) proposed in \cite{fu2021siamese}, and better accuracy of UIM (namely the full model) can be consistently observed.

\begin{table}[t!]

  \renewcommand{\arraystretch}{1.0}
  \vspace{0.1cm}
    \caption{Ablation results, where the meanings of setting abbreviations can be found in Sec. \ref{Abl.1}$\sim$\ref{Abl.5}. The best are stressed in \bf{BOLD}.}\label{tab_ab1}
    \vspace{-8pt}
   \centering
    \footnotesize
    \setlength{\tabcolsep}{1.30mm}
    \begin{tabular}{c|cc|cc|cc|cc}
    \hline
    \multirow{2}{*}{}&\multicolumn{2}{c|}{DAVIS}&
    \multicolumn{2}{c|}{DAVSOD}&\multicolumn{2}{c|}{VOS}&\multicolumn{2}{c}{RDVS}\cr\cline{2-9}
    &$F_{\beta}^{\textrm{max}}\uparrow$&$S_\alpha\uparrow$&$F_{\beta}^{\textrm{max}}\uparrow$&$S_\alpha\uparrow$&$F_{\beta}^{\textrm{max}}\uparrow$&$S_\alpha\uparrow$&$F_{\beta}^{\textrm{max}}\uparrow$&$S_\alpha\uparrow$\cr
    \hline
    A1&0.912&0.924&0.715&0.796&0.785&0.843&0.765&0.843\cr
    \hline
    B1&0.917&0.928&0.733&0.807&0.787&0.850&0.784&0.854\cr
    B2&0.915&0.927&0.743&0.812&0.792&0.850&0.799&0.865\cr
    \hline
    C1&0.914&0.926&0.734&0.805&0.781&0.844&0.783&0.851\cr
    C2&0.919&0.929&0.740&0.811&0.792&0.852&0.797&0.861\cr
    \hline
    D1&0.913&0.926&0.731&0.807&0.783&0.842&0.774&0.850\cr
    D2&0.915&0.928&0.742&0.810&0.792&0.848&0.798&0.859\cr
    D3&0.918&0.928&0.741&0.813&0.785&0.846&0.787&0.855\cr
    D4&0.916&0.927&0.739&0.812&0.786&0.850&0.785&0.857\cr
    D5&0.914&0.926&0.736&0.807&0.785&0.842&0.791&0.855\cr
    \hline
    E1&0.916&0.928&0.738&0.808&0.781&0.842&0.772&0.844\cr
    E2&0.911&0.925&0.735&0.809&0.793&0.851&0.799&0.863\cr
    E3&0.909&0.921&0.728&0.801&0.797&0.855&0.762&0.843\cr
    E4&0.915&0.927&0.730&0.804&0.791&0.848&0.785&0.851\cr
    E5&0.919&0.929&0.740&0.809&0.793&0.852&0.803&0.863\cr
    \hline

    Ours&\bf 0.922& \bf 0.930&\bf 0.754&\bf0.818&\bf0.802&\bf0.858&\bf0.814&\bf0.869\cr
    \hline
    \end{tabular}
    \vspace{-0.4cm}
\end{table}

\subsubsection{HMAPs for refinement} \label{Abl.5}
We first remove the HMAPs (but keep the coarse map supervision) to obtain variant F1. As shown in Table\:\ref{tab_ab2}, it proves the refinement power of HMAPs. Variant F2, on the other hand, switches the location of HMAPs from before RFM to after RFM, proving the validity of refining multi-modal features prior to RFM instead of refining the fused ones. Variant F3 adopts HMAPs to refine all the five layers. It yields degraded performance as deeper features are less redundant to refine than shallower features.

\subsubsection{Hybrid loss for supervision} \label{Abl.6}
To investigate the effectiveness of the hybrid loss function, three variants are conducted in Table\:\ref{tab_ab2}: removing coarse map supervision (G1), removing IoU loss (G2), and removing BCE loss (G3). It can be seen that compared with BCE, ablating IoU loss yields worse effect, whereas the adopted hybrid loss leads to the best results.

\begin{table}[t!]

  \renewcommand{\arraystretch}{1.0}
  \vspace{0.1cm}
    \caption{Ablation results, where the meanings of setting abbreviations can be found in Sec. \ref{Abl.6}$\sim$\ref{Abl.7}. The best are stressed in \bf{BOLD}.}\label{tab_ab2}
    \vspace{-8pt}
   \centering
    \footnotesize
    \setlength{\tabcolsep}{1.30mm}
    \begin{tabular}{c|cc|cc|cc|cc}
    \hline
    \multirow{2}{*}{}&\multicolumn{2}{c|}{DAVIS}&
    \multicolumn{2}{c|}{DAVSOD}&\multicolumn{2}{c|}{VOS}&\multicolumn{2}{c}{RDVS}\cr\cline{2-9}
    &$F_{\beta}^{\textrm{max}}\uparrow$&$S_\alpha\uparrow$&$F_{\beta}^{\textrm{max}}\uparrow$&$S_\alpha\uparrow$&$F_{\beta}^{\textrm{max}}\uparrow$&$S_\alpha\uparrow$&$F_{\beta}^{\textrm{max}}\uparrow$&$S_\alpha\uparrow$\cr
    \hline
    F1&0.918&0.928&0.735&0.804&0.789&0.846&0.794&0.856\cr
    F2&0.916&0.929&0.737&0.809&0.796&0.854&0.792&0.858\cr
    F3&0.920& 0.929&0.736&0.808&0.787&0.844&0.804&0.864\cr
    \hline
    G1&0.919&0.928&0.739&0.811&0.792&0.850&0.793&0.859\cr
    G2&0.914&0.924&0.738&0.813&0.785&0.847&0.770&0.846\cr
    G3&0.915&0.925&0.740&0.809&0.800&0.847&0.787&0.850\cr
    \hline

    Ours&\bf 0.922& \bf 0.930&\bf 0.754&\bf0.818&\bf0.802&\bf0.858&\bf0.814&\bf0.869\cr
    \hline
    \end{tabular}
    \vspace{-0.4cm}
\end{table}

\subsubsection{Computational efficiency} \label{Abl.7}

\begin{table}[t!]

  \renewcommand{\arraystretch}{1.0}
  \vspace{0.1cm}
    \caption{Computational efficiency of DCTNet+, where the speed is represented by inference time (seconds).}\label{ce}
    \vspace{-8pt}
   \centering
    \footnotesize
    \setlength{\tabcolsep}{1.30mm}
    \begin{tabular}{c|c|c|c}
    \hline
    Item&Overall&MAM&RFM \cr
    \hline
    Speed (s)&0.0844&0.0081&0.0325 \cr
    Params (M)&94.6528&0.7415&5.9144 \cr
    \hline
    \end{tabular}
    \vspace{-0.4cm}
\end{table}

The computational efficiency and parameter quantity of DCTNet+, including the overall model, MAM modules, and RFM modules, were tested. The inference time was completed on a desktop equipped with an Intel i5-10500 CPU (3.10GHz), 16G RAM and NVIDIA RTX3060 GPU. We test the inference time of our models over 100 samples (resized to $448 \times 448$). The results are given in Table\:\ref{ce}, from which one can see that MAMs and RFMs together  occupy only $\sim 7\%$ parameters of the overall model.

\section{CONCLUSION}
\label{sec:conclusion}
In this paper, our primary contributions to advance the field of RGB-D VSOD are twofold: the establishment of a novel dataset and a model. Firstly, we create a new dataset for RGB-D VSOD called RDVS with realistic depth. This dataset comprises a variety of scenes with scientific frame-by-frame annotations and is rigorously validated through comprehensive analyses. Furthermore, we present DCTNet+, a specialized model designed for RGB-D VSOD, where RGB is taken as the primary modality, and depth/OF serve as auxiliary modalities. We enhance feature quality and fusion for accurate prediction through two innovative modules: the multi-modal attention module (MAM) and the refinement fusion module (RFM), and also incorporate the universal interaction module (UIM) and holistic multi-modal attentive paths (HMAPs) to improve performance.
We demonstrate the superiority of DCTNet+ through a comprehensive set of experiments conducted on both the newly introduced RDVS and RGB-D video datasets containing pseudo depth maps. Ablation experiments are conducted within this context to showcase the efficacy of each individual module and the positive impact of introducing realistic depth.
With the growing prevalence of RGB-D videos in our daily lives, we believe in the applicable value of RGB-D VSOD and hope our work could inspire more related works/tasks, \emph{e.g.}, instance segmentation, in the future.

\bibliographystyle{IEEEtran}
\bibliography{mybib}

\end{document}